\documentclass{article}

\usepackage{microtype}
\usepackage{graphicx}
\usepackage{subfigure}
\usepackage{booktabs} 
\usepackage{amsmath}
\usepackage{hyperref}

\usepackage[accepted]{icml2025}

\usepackage{amsmath}
\usepackage{amssymb}
\usepackage{mathtools}
\usepackage{amsthm}
\usepackage{float}
\usepackage{multirow}
\usepackage{placeins}
\usepackage{multicol}
\usepackage{cuted}

\usepackage[capitalize,noabbrev]{cleveref}

\theoremstyle{plain}

\theoremstyle{definition}

\theoremstyle{remark}

\usepackage[textsize=tiny]{todonotes}

\icmltitlerunning{GHOST 2.0: Generative High-fidelity One Shot Transfer of Heads}

\begin{document}

\twocolumn[
\icmltitle{GHOST 2.0: \underline{G}enerative \underline{H}igh-fidelity \underline{O}ne \underline{S}hot  \underline{T}ransfer of Heads}



\icmlsetsymbol{equal}{*}

\begin{icmlauthorlist}
\icmlauthor{Alexander Groshev$^{*1}$}{}
\icmlauthor{Anastasiia Iashchenko$^{*1}$}{}
\icmlauthor{Pavel Paramonov$^{*1}$}{}
\icmlauthor{Denis Dimitrov$^{**1, 2}$}{}
\icmlauthor{Andrey Kuznetsov$^{**1, 2}$}{}
\newline \centering $^1$\large{Sber AI} \hspace{0.5cm} $^2$\large{AIRI}

\vspace{20pt}

\end{icmlauthorlist}


\icmlcorrespondingauthor{Andrey Kuznetsov}{kuznetsov@airi.net}
\icmlcorrespondingauthor{Denis Dimitrov}{dimitrov@airi.net}

\icmlkeywords{Head swap, avatars, digital humans, reenactment, image blending, image synthesis}

\vskip 0.1in
]


\begin{figure}[H]
\begin{tabular}{*{3}{@{\hspace{0pt}}c}}
  Source & Target & Result \\
  \includegraphics[width=0.33\columnwidth]{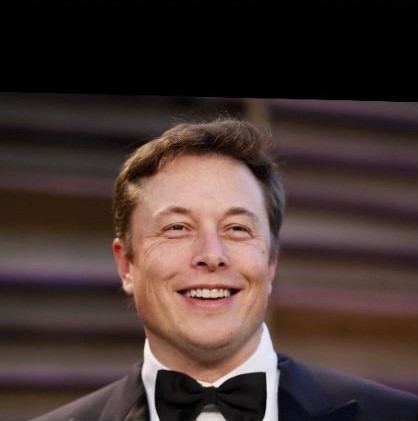}&
  \includegraphics[width=0.33\columnwidth]{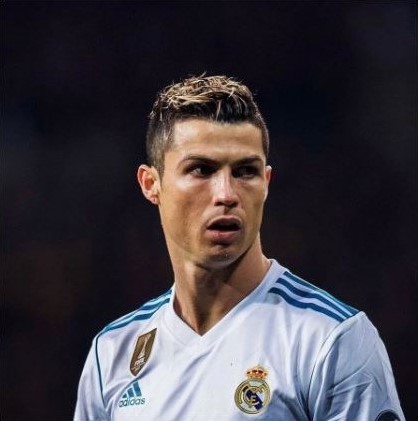}&
  \includegraphics[width=0.33\columnwidth]{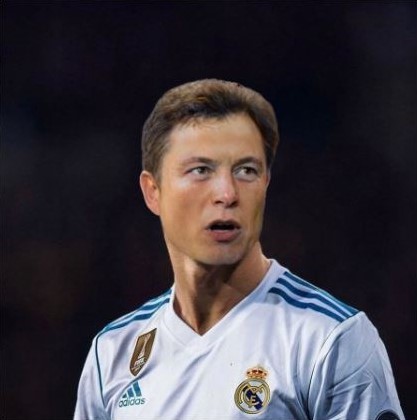} \\
  \includegraphics[width=0.33\columnwidth]{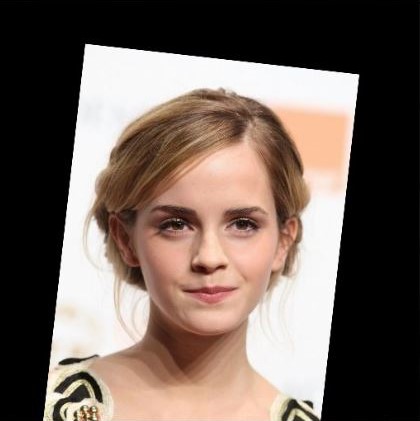}&
  \includegraphics[width=0.33\columnwidth]{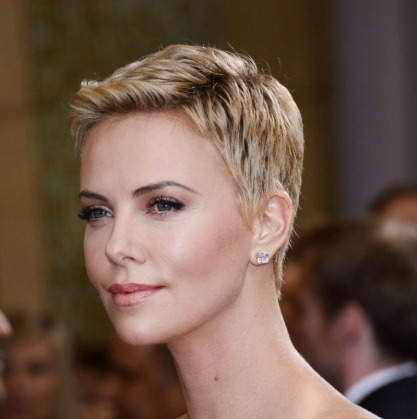}&
  \includegraphics[width=0.33\columnwidth]{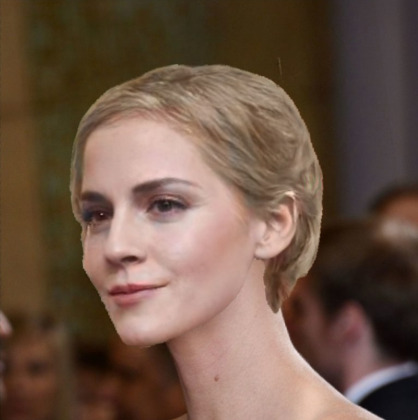} \\
  \includegraphics[width=0.33\columnwidth]{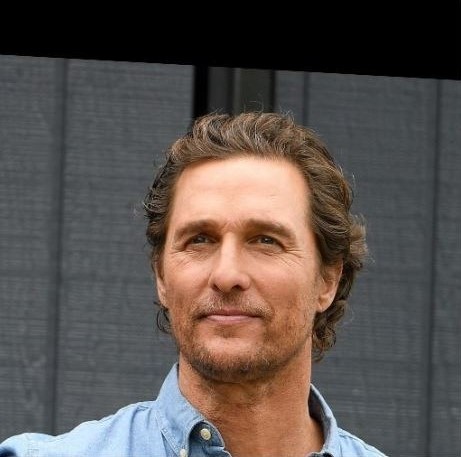}&
  \includegraphics[width=0.33\columnwidth]{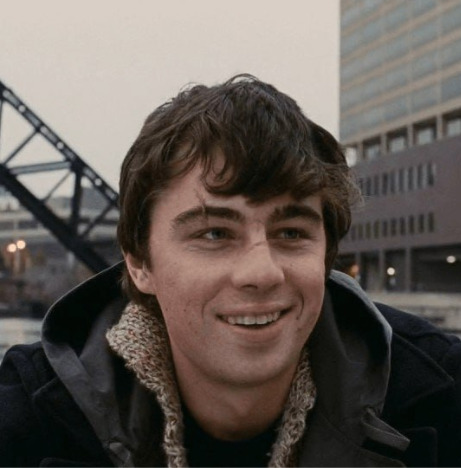}&
  \includegraphics[width=0.33\columnwidth]{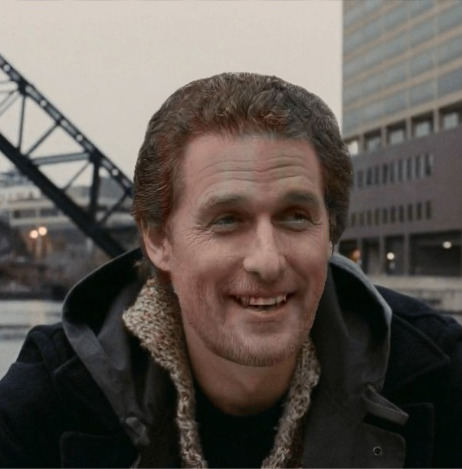} \\  
  \includegraphics[width=0.33\columnwidth]{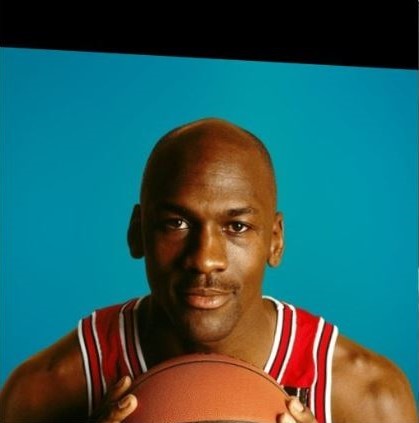}&
  \includegraphics[width=0.33\columnwidth]{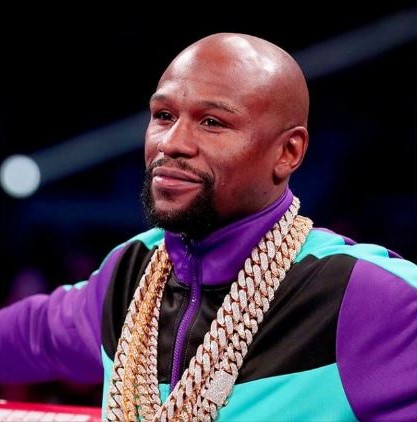}&
  \includegraphics[width=0.33\columnwidth]{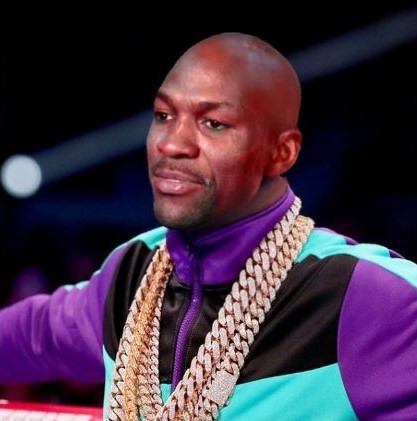} \\ 
  \end{tabular}
  \caption{Results of GHOST 2.0 model on the task of head swap. Head from source image is animated in correspondence with the target motion and blended into target background.}
\end{figure}

\begin{abstract}
While the task of face swapping has recently gained attention in the research community, a related problem of head swapping remains largely unexplored. In addition to skin color transfer, head swap poses extra challenges, such as the need to preserve structural information of the whole head during synthesis and inpaint gaps between swapped head and background. In this paper, we address these concerns with GHOST 2.0, which consists of two problem-specific modules. First, we introduce enhanced Aligner model for head reenactment, which preserves identity information at multiple scales and is robust to extreme pose variations. Secondly, we use a Blender module that seamlessly integrates the reenacted head into the target background by transferring skin color and inpainting mismatched regions. Both modules outperform the baselines on the corresponding tasks, allowing to achieve state-of-the-art results in head swapping. We also tackle complex cases, such as large difference in hair styles of source and target.
\end{abstract}

\section{Introduction}
\label{submission}

\let\thefootnote\relax\footnotetext{$^*$ indicates equal contribution}
\let\thefootnote\relax\footnotetext{$^{**}$ Corresponding authors: Andrey Kuznetsov $<$kuznetsov@airi.net$>$, Denis Dimitrov $<$dimitrov@airi.net$>$}

The use of virtual humans has long passed purely entertainment scope, finding applications in movie and advertisement composition, virtual try-on, deepfake detection, and portrait editing. Head swap, the task of replacing head in the target image with head from the source image, plays an integral role for these use-cases. It requires reenacting source head with the driving one, and seamlessly integrating the result with the target`s background. Active research is conducted on generation and animation of head avatars, mainly using warping, generative adversarial networks (GANs) and diffusion models \cite{Siarohin_2019_NeurIPS,  Drobyshev22MP, zhang2022metaportrait, kirschstein2023diffusionavatars}. However, the problem of blending the generated head with the surrounding environment remains largely unaddressed. 

Recently, a noticeable progress in a related task of face swapping has been made \cite{simswap2020, hififace2021, ghost2022, zhu2021megafs, zhao2023diffswap}. The task requires source identity preservation and reenactment only within facial region. Since information about head shape and hair is omitted, face swap is a less complex problem than swapping of the whole head for several reasons. First, face recognition models \cite{cao2018vggface2, deng2019arcface, ghostfacenet2023}, usually used to control identity transfer, are trained on narrow face crops. Thus, they cannot be used to embed identity information about the whole head, requiring consideration of other models for this purpose. Secondly, head generation also involves generation of hair, which is a complex task due to its high-frequency texture and diverse styles. Finally, variations in face shapes are usually smaller than variations in head forms. This implies potentially larger regions for inpainting as a result of head replacement. And, similarly to face swap, the skin color of target should be carefully transferred to the generated head for a realistic result.


Currently there are only few papers that address the task of head swap. The first steps were taken by DeepFaceLab \cite{perov2020deepfacelab}, which   requires finetuning for each source identity. Next work, StylePoseGAN \cite{sarkar2021style}, tends to modify background and skin color of the target. A seminal work that partially tackles the aforementioned issues is HeSer \cite{shu2022few}. It splits the generation process into head reenactment stage and reference-based blending stage. This work was followed by diffusion-based approaches \citep{baliah2025realistic, han2023generalist}. However, latest methods still face issues concerning quality of head generation and color transfer.

In this paper we present GHOST 2.0, a one-shot high-fidelity framework for head swap. It consists of two modules: Aligner  for head reenactment, and Blender for natural inpainting of the result into the target background. Aligner includes a set of encoders to obtain information on source identity and target motion, which is then used to condition StyleGAN-based generator to obtained aligned head. The choice of architecture ensures good in-the-wild performance, supporting high-quality generation even in extreme poses. Blender creates references for head color transfer and background inpainting, and uses them to stitch the generated head with the background via blending UNet \cite{ronneberger2015u}. For color transfer, the composition of reference is based on correlation between respective head parts of generated and driving head. Background reference is determined by masking the potential gaps arising due to differences in head shape. These references are used to condition UNet which outputs final result. 

Our contributions are the following:
\begin{itemize}
    \item We introduce a new model for head reenactment that surpasses competitors on a range of metrics. In contrast to previous works that focus on face region only, it generates full human head, accounting both for low-frequency and high-frequency details and preserving identity at different scales. Moreover, when trained on the same dataset as other methods, it gives superior results on generation in extreme poses.  
    \item We increase quality of final generations by improving robustness of blending module to corner-case scenarios. Specifically, these include the situation when hair styles of source and target are extremely different, which previously led to inconsistent hair transfer. Additionally, we consider the case when target head lacks color references for the source one, resulting in poor color transfer. Finally, we enhance background inpainting to achieve more seamless blending.
    \item We trained a new segmentation model specifically for the head swap task. Unlike existing segmentation models, we have annotated the data in such a way as to separate beard and facial hair into a separate class, which is necessary for correct color transfer.
\end{itemize}

\section{Related Work}

\paragraph{Face swap}
There are a large number of face-swapping methods. Conceptually, they can be divided into several different groups. Methods from the first group \cite{simswap2020, hififace2021, ghost2022} extract the identity vector and some other features of the source face and use a generative model to blend them with the attributes of the target. Often, such models rely on the ArcFace \cite{deng2019arcface} model and a 3D shape-aware identity extractor, which allows encoding the 3D geometry of the face.
There are also approaches \cite{zhu2021megafs} based on StyleGAN2 \cite{Karras2019stylegan2}. They propose inverting the source and driving images into the latent space and feeding them into the StyleGAN2 generator to perform the swap. This approach allows for higher-resolution results but is sensitive to input data and does not perform well on strong rotations or small details of images.
With the development of diffusion models, approaches to face replacement using this method have emerged \cite{zhao2023diffswap, chen2024hifivfshighfidelityvideo}. Diffusion models allow for high-quality results, but they are typically slow and require significant computational power and sufficient VRAM.

\paragraph{Head swap}
The task of head swap is covered by a limited number of works. DeepFaceLab \cite{perov2020deepfacelab} is the first approach enabling this capability. However, it requires a large amount of source data for training, and poorly performs color transfer and fusion of generated head with the background. StylePoseGAN \cite{sarkar2021style} performs head swap by conditioning StyleGAN \cite{Karras2019stylegan2} on pose and combined texture map, with body parts taken from target and head — from source. 
Still, it tends to modify the background and skin color of the target. HeSer \cite{shu2022few} tackles these issues by designing a separate module for each task. First, it uses head reenactment model based on \cite{Burkov_2020_CVPR} to align source head with target in pose and expression. In the second stage a reference is created for skin color transfer based on correlation between pixels from the same head parts. Together with background inpainting prior, it is used to condition blending UNet \cite{ronneberger2015u} which fuses the head with background. While this method outperforms the competitors, it suffers from identity leakage of target and is unable to color head parts present in source image but absent in the driving one.  
While previous methods are based on GANs \citep{goodfellow2014generative}, there have been attempts \citep{baliah2025realistic, han2023generalist, wang2022hs} to use diffusion models instead \citep{ho2020denoising}. However, these approaches face issues of pose controllability, preservation of target skin color and overall realism of generated head.

\vspace{-8pt}
\paragraph{Head reenactment} 
Head reenactment methods can be generally categorized as either warping-based \cite{Siarohin_2019_NeurIPS, zakharov2020fast, Drobyshev22MP, zhang2022metaportrait, wang2022latent} or reconstruction-based \cite{zielonka2022insta, NEURIPS2023_937ae0e8, qian2024gaussianavatars, chu2024gpavatar, deng2024portrait4d}. Warping-based approaches utilize motion and facial expression descriptors of the target to deform source image. These descriptors can be based on keypoints \cite{Siarohin_2019_NeurIPS, zakharov2020fast}, blendshapes from parametric models \cite{ren2021pirenderer, yin2022styleheat} or latent representations. The latter usually achieves better expressiveness, however, it requires careful disentanglemet of motion from the appearance of the target. This can be achieved via the use of special losses \cite{Drobyshev22MP, drobyshev2024emoportraits, Pang_2023_CVPR} or additional regularization embedded into the architecture \cite{Pang_2023_CVPR}. However, warping-based approaches generally perform well only if the difference between source and target poses is small.
Reconstruction-based methods \cite{zielonka2022insta, NEURIPS2023_937ae0e8, qian2024gaussianavatars, chu2024gpavatar, deng2024portrait4d} construct latent model of source head, and therefore are robust to larger pose deviations. These methods often utilize implicit representations such as TriPlanes \cite{ma2023otavatar, ye2024real3dportrait} and NeRF \cite{zielonka2022insta, zheng2022imavatar, bai2023learning}, or explicit  ones, such as voxels \cite{xu2023avatarmav}, point clouds \cite{zheng2023pointavatar} and meshes \cite{Khakhulin2022ROME, grassal2022neural}, with particularly photorealistic results achieved recently with Gaussian splatting \cite{qian2024gaussianavatars, giebenhain2024npga}. However, reconstruction with these approaches requires an additional step of per-frame estimation of camera parameters, which increases runtime. Also, due to high computational cost of rendering, the resolution of output images does not exceed $256 \times 256$ and upsampling to higher resolutions is performed by an additional network.

\section{Approach}
Our pipeline consists of two modules, Aligner and Blender. Aligner is used to perform cross-reenactment by transferring target motion to source head. Several encoders embed relevant information from input images at different scales, which is then fused in decoder network. Positionally aligning both heads allows to perform further blending. It is based on identifying regions that require inpainting, and construction of color references for them. Color reference for head is obtained via correlation learning, while for background we use LaMa inpainting network \cite{suvorov2021resolution}. They are supplied to UNet network,  which performs final blending of reenacted head into the target background.
\begin{figure}[H]
\vskip -0.2in
\begin{center}
\centerline{\includegraphics[width=\columnwidth]{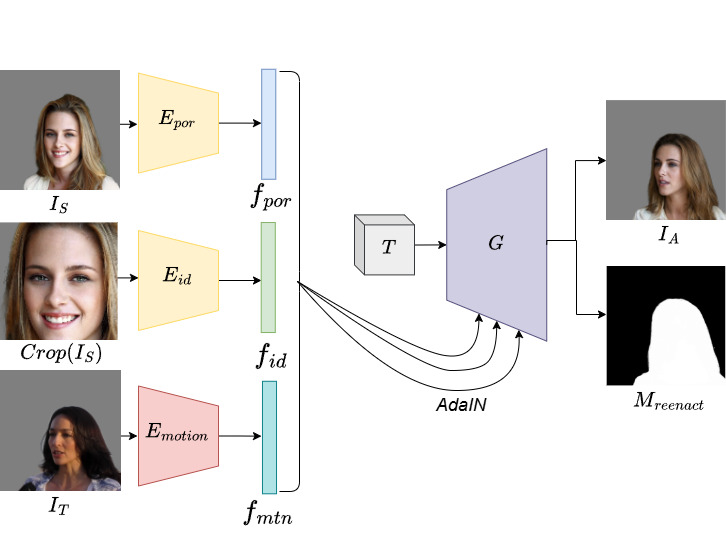}}
\caption{Aligner architecture. Two appearance encoders $E_{por}$ and $E_{id}$ take images of source head $I_S$ and face $Crop(I_S)$ and produce embeddings $f_{por}$ and $f_{id}$. Motion encoder $E_{motion}$ is used to obtain respective embedding $f_{mtn}$ from the target image $I_T$. The embeddings $f_{por}$, $f_{id}$ and $f_{mtn}$ are concatenated and used to condition generator $G$ via AdaIN \cite{huang2017adain} layers. The generator takes a learnable tensor $T$ as input and outputs reenacted head $I_A$ and binary mask $M_{reenact}$.}
\label{fig:aligner}
\end{center}
\vskip -0.2in
\end{figure}

\subsection{Aligner} 
As we target model usage for in-the-wild scenario, we have chosen the reconstruction approach to face reenactment to increase robustness to large pose variations. We decided to use a simple architecture for a faster and more lightweight solution. Thus it is based on 2D instead of 3D volumetric representations  to remove the need for camera estimation, rendering and additional upsampling to higher resolution. However, in principle any reenactment model can be used at this stage, provided it reconstructs the whole head.

\paragraph{Aligner architecture}
Aligner module, based on \cite{shu2022few}, is illustrated in Fig. \ref{fig:aligner}. It consists of a set of encoders to embed relevant information from source and target images, which is then passed to condition generator. Two encoders, $E_{id}$ and $E_{por}$, are used to encode identity of the source at multiple scales. Local information is encoded by $E_{id}$, a pretrained state-of-the-art face recognition network \cite{deng2019arcface}. It takes central face crops $Crop(I_S)$ as input and outputs face embedding $f_{id} \in \mathbb{R}^{512}$. Global information, including hair and head shape, is extracted by $E_{por}$, which takes full source image and outputs $f_{por} \in \mathbb{R}^{512}$. Such combination of appearance encoders allows to obtain more fine-grained head reconstruction, while  paying special attention to more discriminative facial features. 

To embed driving head pose and facial expression, we use motion encoder $E_{motion}$. Given full augmented target image $I_T$, it produces motion embedding $f_{mtn} \in \mathbb{R}^{256}$ . We address disentanglement of motion from appearance by stretching $I_T$ independently along horizontal and vertical axes before supplying it to $E_{motion}$. In this way, we change identity of the person, while preserving head pose and expression.

The obtained descriptors $f_{id}$, $f_{por}$ and $f_{mtn}$ are concatenated and used to condition StyleGAN-like generator $G$ via AdaIN \cite{huang2017adain}. It takes learnable tensor $T \in \mathbb{R}^{512 \times 4 \times 4}$ as input and passes it through a series of convolutional blocks to obtain final image $I_A$, injecting the appearance and motion information at each step. To stabilize training, the generator outputs binary head mask $M_{reenact}$ along with reenacted head.

\begin{figure}[H]
  \centering
  \begin{tabular}{*{5}{@{\hspace{0pt}}c}}
  Source & Target & $I_{exp}$ & $I_{pose}$ & $I_{full}$ \\
    \includegraphics[width=0.2\columnwidth]{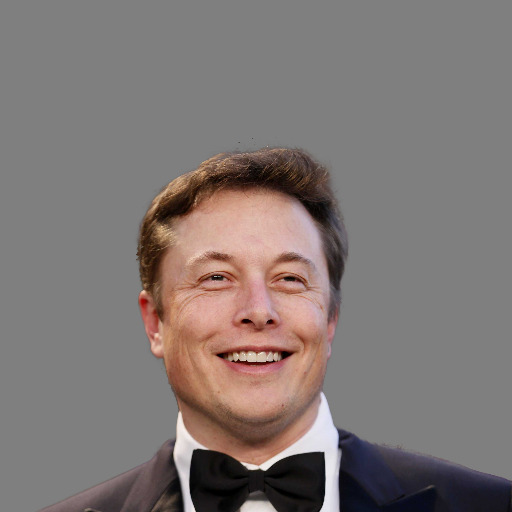}&
    \includegraphics[width=0.2\columnwidth]{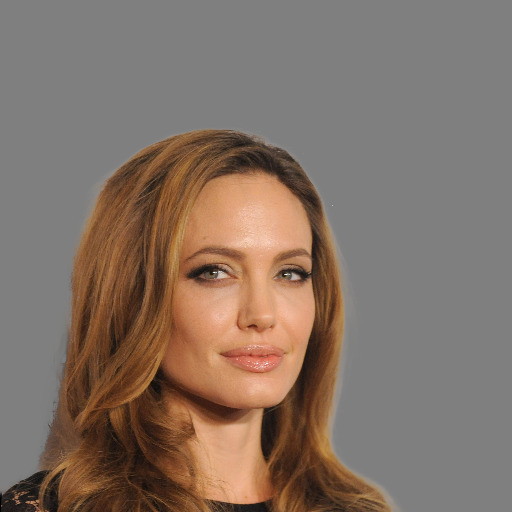}&
    \includegraphics[width=0.2\columnwidth]{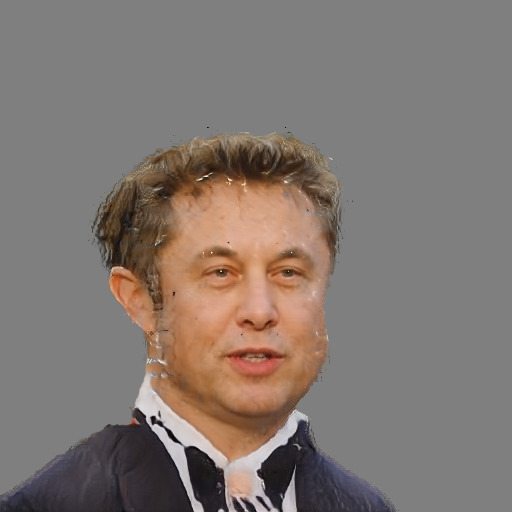}&
    \includegraphics[width=0.2\columnwidth]{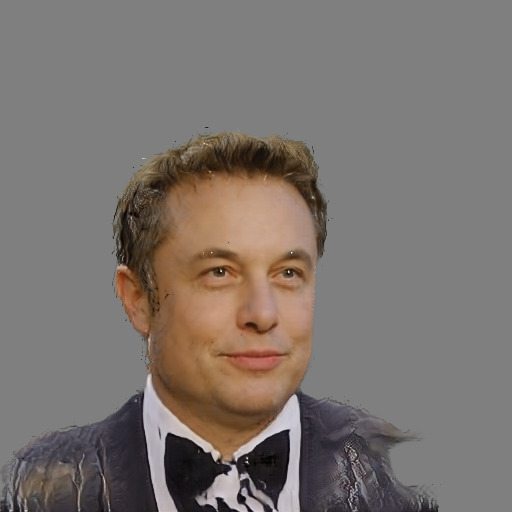}&
    \includegraphics[width=0.2\columnwidth]{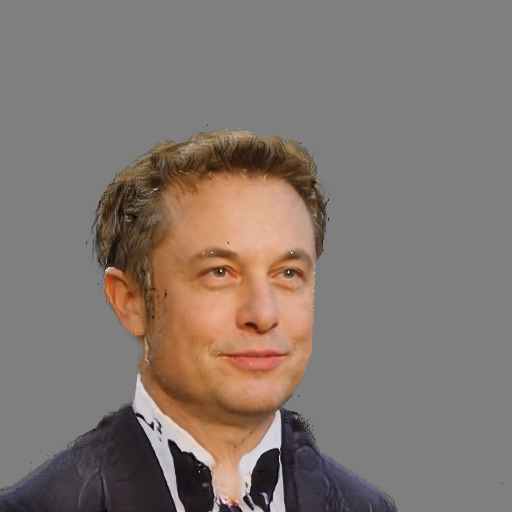}\\
  \end{tabular}
  \caption{Ablation results on representations learned by pose $E_{pose}$ and expression $E_{exp}$ encoders}
  \label{fig:enc_abl}
\end{figure}

\paragraph{Refined motion encoder}
The original Aligner architecture of HeSer \cite{shu2022few} included separate pose $E_{pose}$ and expression $E_{exp}$ encoders. However, we encountered the problem of target identity leakage with this design, resulting in the mixing of target and source identitties in the output image $I_A$ . To obtain more insights into the problem, we conducted ablation study to learn which information is embedded by each encoder (fig. \ref{fig:enc_abl}). We used embedding of canonical pose as output of $E_{pose}$, and allowed $E_{exp}$ to obtain expression embedding from target image, resulting in generation $I_{pose}$. In this case, head is still generated in pose of target, not in canonical one. It is overall quite similar to full result $I_{full}$, except for difference in skin color which is now closer to the source one. On the other hand, by using canonical expression and image-based pose representation we obtain final generation $I_{exp}$ with almost canonical expression and pose. This ablation indicates that $E_{exp}$ learns both expression and pose embeddings, while $E_{pose}$ mainly transfers appearance of the target.

Next we decided to regularize each encoder to split the motion information into meaningful separate embeddings. For this we tested different approaches by 1) decreasing size of encoders, 2) using disentanglement and cycle losses as in \cite{Drobyshev22MP, Pang_2023_CVPR}, 3) using pretrained encoders from \cite{DECA:Siggraph2021, Drobyshev22MP}. However, these experiments resulted in inferior performance compared to the baseline of single motion encoder $E_{motion}$, which embeds both pose and expression into single vector $f_{mtn} \in \mathbb{R}^{256}$. Relevant ablation is shown in section results. 

\vspace{-10pt}
\paragraph{Training losses} To train Aligner, we use hinge adversarial loss $\mathcal{L}_{adv}$, feature-matching  loss $\mathcal{L}_{FM}$ \cite{salimans2016improved},  L1 reconstruction loss $\mathcal{L}_{L1}$, VGG-based  perceptual loss  $\mathcal{L}_{perc}^{VGG}$ \cite{simonyan2014very} and dice loss  $\mathcal{L}_{dice}$ \cite{sudre2017generalised}. To improve source identity preservation, we introduce cosine $\mathcal{L}_{cos}^{ID}$ and perceptual $\mathcal{L}_{perc}^{ID}$ losses comparing embeddings and feature maps of $I_A$ and $I_S$ obtained with our pretrained encoder $E_{id}$.
We also noticed that with the new combined motion encoder $E_{motion}$ expressiveness decreases, and the closure of mouth and eyelids does not follow the target. We introduce additional emotion loss ${L}_{emo}$ \cite{EMOCA:CVPR:2021} and keypoint closure loss that compares distance between lips and eyelids for generated and driving heads:

\begin{equation}
    \mathcal{L}_{kpt} = \sum_{k_i^{lower} \in K_{lower}} |k_i^{lower} - k_i^{upper}|
\end{equation}

where $K_{lower}$ is the set of keypoints on lower eyelid or lip, and $k_i^{lower}$ and $k_i^{upper}$ are symmetric keypoints on lower and upper eyelid or lip, respectively. Since reenactment quality is also influenced by such subtle factors as gaze direction, we also include perceptual gaze loss as in \cite{Drobyshev22MP} starting from 1000 epochs. Additional details on losses are given in supplementary material.

\begin{figure*}[ht]
\vskip 0.2in
\begin{center}
\centerline{\includegraphics[width=2\columnwidth]{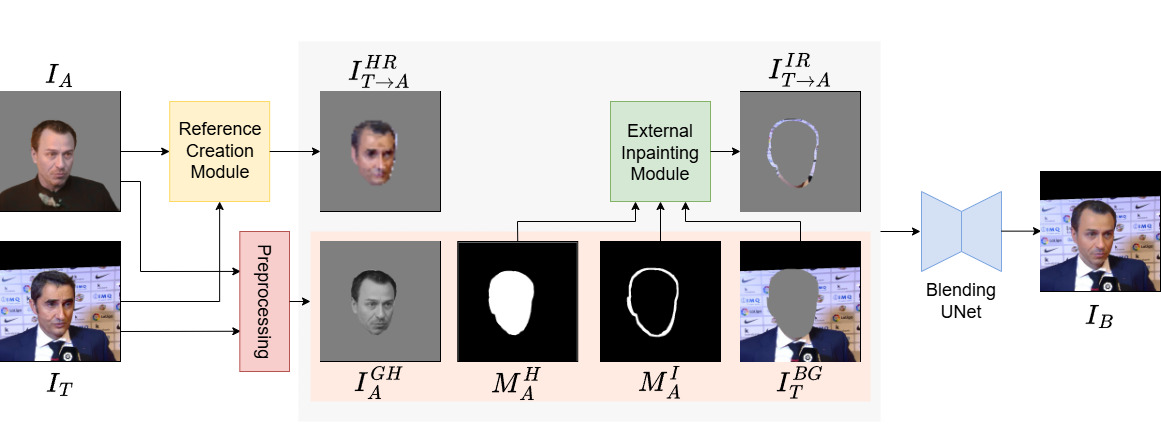}}
\caption{Blender architecture. Reenacted $I_A$ and target $I_T$ images are preprocessed to obtain grayscale animated head $I_A^{GH}$, target background $I_T^{BG}$ and masks $M_A^H$ and $M_A^I$ defining head and background inpainting regions. Also, $I_A$ and $I_T$ are passed to Reference Creation module to obtain head color reference $I_{T\rightarrow A}^{HR}$ for color transfer. The background reference $I_{T\rightarrow A}^{IR}$ is obtained via External Inpainting module \cite{suvorov2021resolution}. These inputs are passed to Blending UNet to achieve final result $I_B$.}
\label{aligner}
\end{center}
\end{figure*}

\subsection{Blender}
\subsubsection{Preliminary}
We base our Blender on the corresponding module from \cite{shu2022few}. It involves data preprocessing, color reference creation, and blending steps. 

\begin{figure}[H]
\begin{center}
\centerline{\includegraphics[width=\columnwidth]{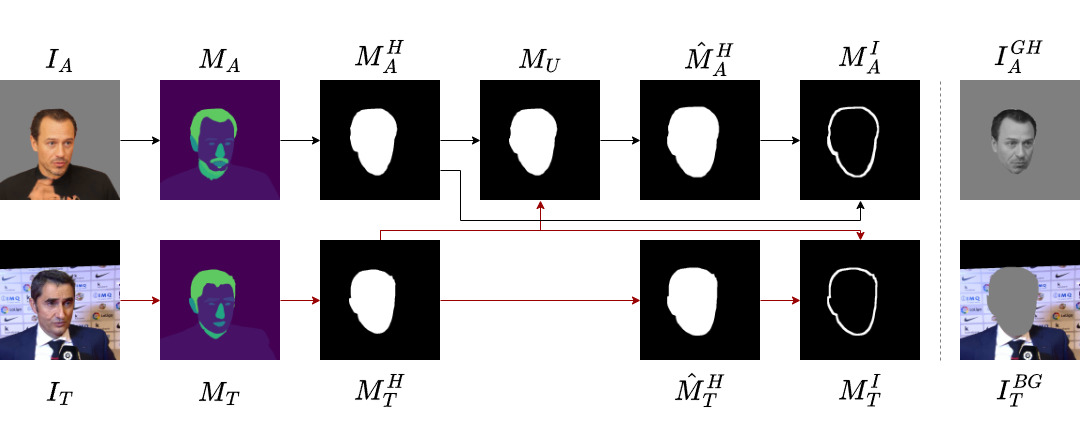}}
\caption{Masks obtained in data preprocessing stage}
\label{aligner}
\end{center}
\vskip -0.2in
\end{figure}

\paragraph{Data preprocessing}
Data preprocessing stage prepares inputs for color transfer and background inpainting. In particular, the first problem is viewed as a problem of re-coloring gray-scale reenacted head with colors of the target head. For this, segmentation masks $M_A$ of the head regions and a binary head mask $M_A^H$ are obtained from reenacted image $I_A$. Similarly, target segmentation $M_T$ and head $M_T^H$ are obtained from target image $I_T$. Then gray-scale image of the head to be colored is obtained as $I_{A}^{GH} = Gray(I_A *M_A^H)$

Additionally, we need to define regions that need inpainting due to differences in head shape between source and target. For this, union $M_U$ of reenacted $M_A^H$ and target $M_T^H$ head masks is dilated to an enlarged head mask $\hat{M}_A^H$. Then the area that requires inpainting is denoted as $M_A^I=\hat{M}_A^H-M_A^H$. The region to serve as color reference for background inpainting is then defined as $M_T^I=\hat{M}_T^H-M_T^H$, where $\hat{M}_T^H$ is dilated target head mask $M_T^H$. Finally, background without head can be obtained as  $I_T^{BG}=I_T*(1-\hat{M}_A^H)$.

Additionally, we introduce augmentation of inpainting mask  $M_A^I$ to enhance inpainting of large mismatched regions.

\begin{figure}[H]
\begin{center}
\centerline{\includegraphics[width=\columnwidth]{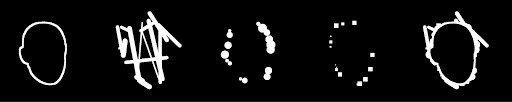}}
\caption{Examples of inpainting mask  $M_A^I$ augmentations}
\label{aligner}
\end{center}
\vskip -0.2in
\end{figure}

\paragraph{Color reference creation}
The next step is to provide color references for background inpainting and head color transfer. Creation of these references is based on learning the correlation between corresponding semantic regions of input $I_A$ and target $I_T$ images. During training of corresponding Reference Creation (RC) module, the same image serves as both input $I_A$ and target $I_T$. To prevent the network from merely copying pixels from the same position, random color augmentation $C'$ is applied to $I_A$ and random horizontal flip $F$ is applied to $I_T$. Correlation learning takes place in the latent space based on the representation obtained with Feature Pyramid Network (FPN) \cite{lin2017feature} for $I_A$ and $I_T$:
\begin{equation}
    f_A=FPN(C'(I_A))
\end{equation}
\begin{equation}
    f_T=FPN(F(I_T))
\end{equation}


Next, correlation is calculated between features $f_A$ and $f_T$ for each spatial location. It is used to weight pixels during resampling in the following step. For each semantic region $r \in \{\text{face, ears, eyes, brows, nose, lips, mouth, teeth, beard, hair,} \newline \text{glasses, hat, headphones, earrings} \}$ a correlation matrix $\Gamma^r \in \mathbb{R}^{N^r_A \times N^r_T}$ is computed, where $N^r_A$ and $N^r_T$ are numbers of pixels in region $r$ in input $I_A$ and target $I_T$ images. Thus, each element $\Gamma^r(u, v)$ is calculated as:
\begin{equation}
    \Gamma^r(u, v) = \frac{\bar{f}_A^r(u)^T\bar{f}_T^r(v)}{\|\bar{f}_A^r(u)\| \|\bar{f}_T^r(v)\|}, \hspace{3pt} u \in M^r_A, \hspace{3pt} v \in M_T^r
\end{equation}
where $\bar{f}_A^r(u)$ and $\bar{f}_T^r(v)$ are channelwise centralized features $f_A$ and $f_T$ at locations $u$ and $v$. $\Gamma^r(u, v)$ is then normalized via softmax and used to determine contribution of each pixel $v$ in corresponding region of $I_T$ to the color of pixel $u$ in the head reference image $I^r_{T \rightarrow A}(u)$:
\begin{equation}
    I^r_{T \rightarrow A}(u) = \sum_{u \in M^r_T}^{} \text{softmax}_v (\Gamma^r(u, v)/ \tau) \cdot I_T(v), \hspace{3pt} u \in M^r_A
\end{equation}
where $\tau$ is temperature coefficient. Overall, computation of head color reference $I^{HR}_{T \rightarrow A}$ is based on the following inputs: 
\begin{equation}
    I^{HR}_{T \rightarrow A} = RC(f_A, M_A, f_T, M_T, I_T*M_T^H)
\end{equation}

However, during inference a label mismatch can occur if an area is present in the reenacted image $I_A$ but not in the target one $I_T$. In this case, such face region is not colored by the blending UNet, since no color reference for it is given. To remedy this problem, we propose to assign colors from other semantically relevant areas of $I_T$. In case no such area is found (for instance hat is present  in $I_A$ but not in $I_T$) or is too small to provide robust reference, we copy the colors from the original image $I_A$.

To create color inpainting reference $I^{IR}_{T \rightarrow A}$, we use LaMa model \cite{suvorov2021resolution}, conditioned on inpainting mask $M^I_T$. We found this solution to provide superior quality compared to creation with RC module. 

\paragraph{Blending UNet}
Blending of reenacted head into target background is performed by Blending UNet $B$ submodule. As input it takes concatenated head $I^{HR}_{T \rightarrow A}$ and background inpainting $I^{IR}_{T \rightarrow A}$ references, head mask $M_A^H$, background $I^{BG}_T$, inpainting mask $M_A^I$ and gray-scale head $I^{GH}_A$ and outputs final image $I_B$:
\begin{equation}
    I_B = B(I^{HR}_{T \rightarrow A}, I^{IR}_{T \rightarrow A}, M_A^H, I^{BG}_T, M_A^I, I^{GH}_A)
\end{equation}

The UNet is trained with standard adversarial $\mathcal{L}_{adv}$, reconstruction $\mathcal{L}_{L1}$ and perceptual losses $\mathcal{L}_{perc}^{VGG}$. Additionally, to learn a meaningful correlation matrix $\Gamma^r$, cycle consistency loss is used:
\begin{equation}
    L_c = \lambda_c \|I_{T \rightarrow A \rightarrow T} - I_T\|_1
\end{equation}
where $I_{T \rightarrow A \rightarrow T}^k(u) = \sum_{v \in M^k_A}^{} \text{softmax}_v (\Gamma^k(u, v)/ \tau) \cdot I_{T \rightarrow A}(v), \hspace{3pt} u \in M^k_T$. Additional regularization is performed by calculation of cycle loss with another target image $I_{T'}$ having the same identity as $I_A$: 
\begin{equation}
    L_{c'} = \lambda_{c} \|I_{T' \rightarrow A \rightarrow T'} - I_T\|_1
\end{equation}
To make head color reference more similar to source image, regularization loss is used:
\begin{equation}
    L_{reg} = \lambda_{reg} \|M_A^H\cdot\left((grayscale(I_A) - grayscale(I^{HR}_{T\rightarrow A})\right)\|_1
\end{equation}
More details on corresponding hyperparameters are given in supplementary.

\begin{figure}[H]
  \centering
  \begin{tabular}{*{5}{@{\hspace{0pt}}c}}
  $I_A$ & $I_{ext}$ & $\widehat{I_{A}}$ & $I_B$ & $\widehat{I_B}$ \\
  \includegraphics[width=0.2\columnwidth]{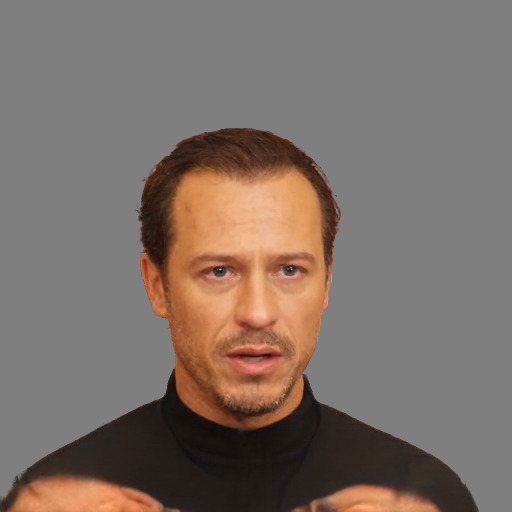}&
    \includegraphics[width=0.2\columnwidth]{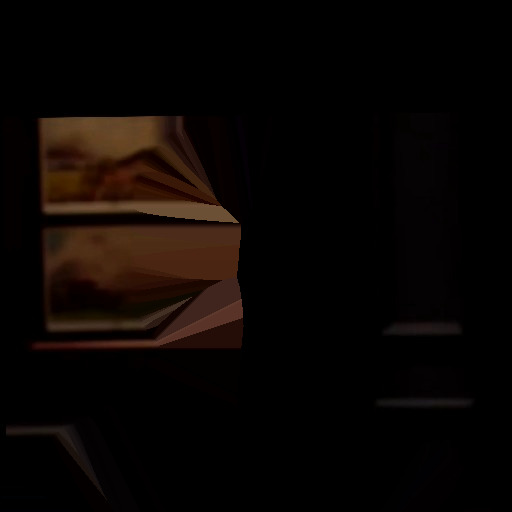}&
    \includegraphics[width=0.2\columnwidth]{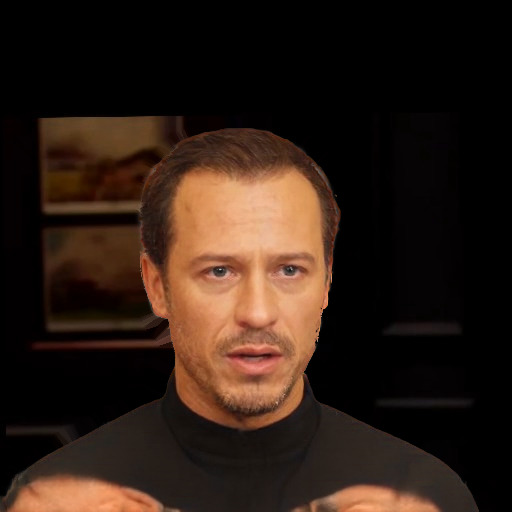}&
    \includegraphics[width=0.2\columnwidth]{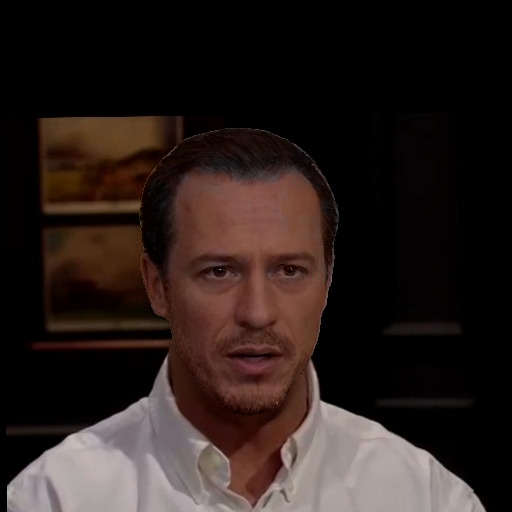}&
    \includegraphics[width=0.2\columnwidth]{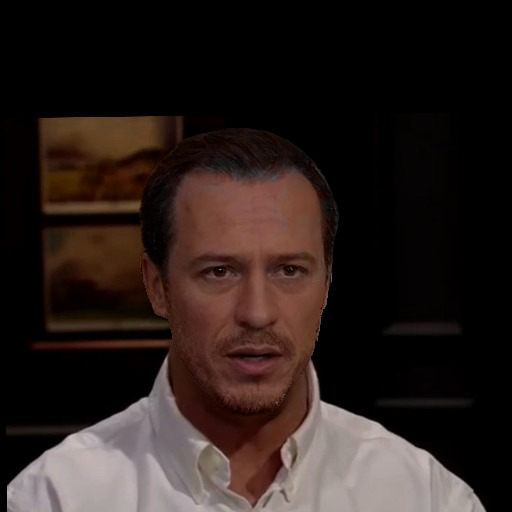}\\
  \end{tabular}
  \caption{Improved blending of hair. $\widehat{I_B}$ and $I_B$ are final results with and without background extrapolation}
  \label{fig:blending_of_hair}
\end{figure}

\paragraph{Improved blending of hair}

We also implement additional step on Aligner output to refine blending of hair in the resulting image $I_B$. We utilize soft portrait masks $M_{soft}$ to segment hair area. They provide better segmentation results than hard ones due to the uncertainty of edge estimation of the hair region. However, due to such choice of masking blending UNet recognizes soft mask areas to belong to the head region and does not extrapolate background to them, resulting in a visible border. To remedy this problem, we create additional image $I_{ext}$, where the background is extrapolated over the head mask $M^H_A$, as shown in fig. \ref{fig:blending_of_hair}. Then we obtain refined animated portrait $\widehat{I_A}$ by blending extrapolated background $I_{ext}$ with Aligner output based on matting mask $M_{soft}$: $\widehat{I_A} = M_{soft} * I_A + (1- M_{soft}) * I_{ext}$.

\paragraph{Head transfer on real data}

\begin{figure}[ht]
\begin{center}
\centerline{\includegraphics[width=\columnwidth]{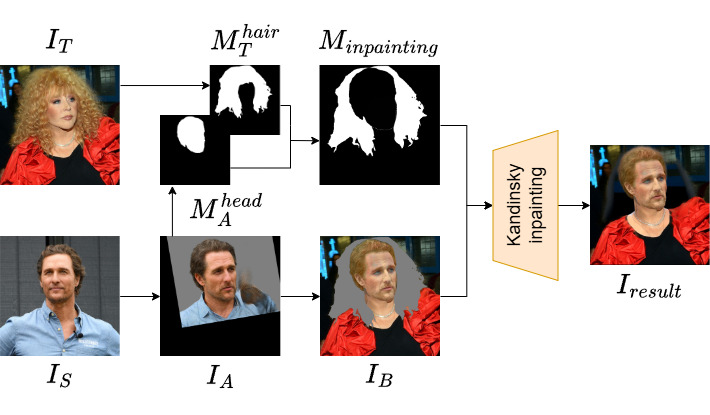}}
\caption{Post-processing of blended image. Given target $I_T$ and reenacted $I_A$ images, we obtain masks of reenacted head $M_A^{head}$ and target hair $M_T^{hair}$. We substract the first mask from the second one to obtain area for inpanting $M_{inpainting}$, which is filled by Kandinsky model \cite{kandinsky2.2}}
\label{post-inpainting}
\end{center}
\vskip -0.2in
\end{figure}

When transferring to real data, we must take into account that our model works within cropped images. In contrast to the face swap problem, where the person's face is always inside the generated region, in the head swap problem the hair can extend beyond the boundaries of the considered area. In this case, we need to remove excess hair from the image. For this purpose, we propose a post-processing blending algorithm using Kandinsky diffusion model \cite{kandinsky2.2}. We obtain the final image by inpaiting the region masked by $M_{inpainting}$ for UNet output $I_B$, as shown in fig. \ref{post-inpainting}. It should be noted that this step is optional and is applied when person in the target image has hair that significantly differs from hair of source.

\begin{figure}[H]
  \centering
  \begin{tabular}{*{3}{@{\hspace{0pt}}c}}
  Image & BiSeNet & Ours\\
  \includegraphics[width=0.33\columnwidth]{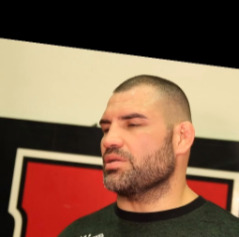}&
    \includegraphics[width=0.33\columnwidth]{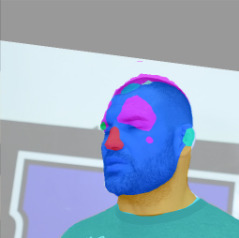}&
    \includegraphics[width=0.33\columnwidth]{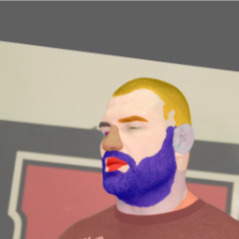} \\
    \includegraphics[width=0.33\columnwidth]{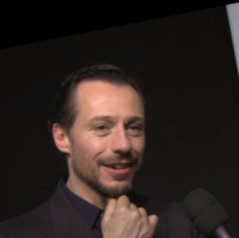}&
    \includegraphics[width=0.33\columnwidth]{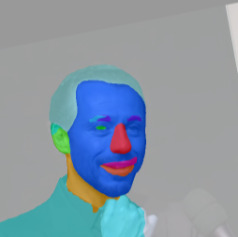}&
    \includegraphics[width=0.33\columnwidth]{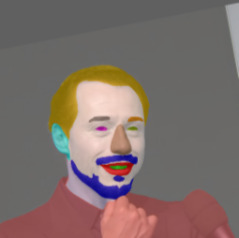} \\
    \includegraphics[width=0.33\columnwidth]{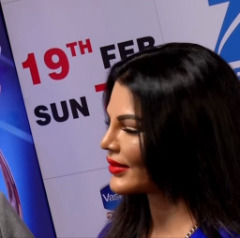}&
    \includegraphics[width=0.33\columnwidth]{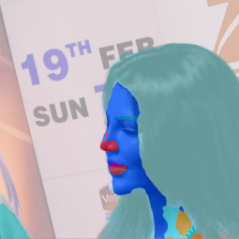}&
    \includegraphics[width=0.33\columnwidth]{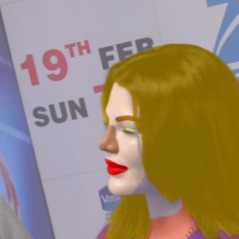} \\
    \includegraphics[width=0.33\columnwidth]{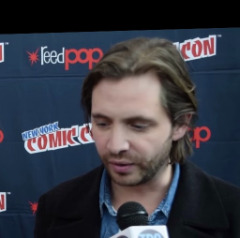}&
    \includegraphics[width=0.33\columnwidth]{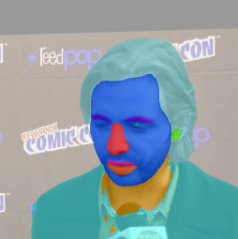}&
    \includegraphics[width=0.33\columnwidth]{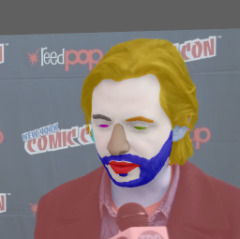} \\
  \end{tabular}
  \caption{Qualitative comparison of segmentation models}
  \label{segmentation}
\end{figure}

\subsection{Segmentation}
For our model to work, it is necessary to have a high-quality segmentation model—it will be used in the Blender module, as well as during preprocessing stages to select a person's head. There are two main requirements for the model: it must be able to segment hairstyles and facial hair as separate classes, and it must be additive, meaning it should segment in such a way that we can combine the segments to obtain a complete head. Additionally, each region should be homogeneous in color—for example, the 'beard' class should not overlap with the 'skin' class, as these regions will be used later for color transfer. To train the model, a dataset of 20,000 FullHD images was collected and annotated \cite{EasyPortrait}, and a segmentation model was trained based on it. We settled on Segformer-B5 \cite{xie2021segformer} as the model for face parsing and segmentation. As a result, the model can segment the following classes: \textit{background}, \textit{person}, \textit{skin}, \textit{left brow}, \textit{right brow}, \textit{left eye}, \textit{right eye}, \textit{mouth}, \textit{teeth}, \textit{lips}, \textit{left ear}, \textit{right ear}, \textit{nose}, \textit{neck}, \textit{beard}, \textit{hair}, \textit{hat}, \textit{headphone}, \textit{glasses}, \textit{earring}. Fig. \ref{segmentation} shows a visual comparison of our model and BiSeNet \cite{yu2018bisenet}.

\section{Experiments}
\subsection{Experiment Setup}

\paragraph{Dataset}
We use VoxCeleb2 \cite{chung2018voxceleb2} dataset to train and evaluate our model at $512 \times 512$ and $256 \times 256$ resolutions. Since some videos originally have low quality, we filter data using image quality assessment methods \cite{Su_2020_CVPR, wang2023exploring}, leaving approximately 70\% of the original dataset. We additionally preprocess the data by cropping face and head regions and calculating keypoints with \citet{DECA:Siggraph2021}. Our final train and test sets include respectively 135500 and 5500 videos. The split is made so as to avoid intersection between train and test identities. During training, we sample source and target from the same video, while during inference they can feature different identities. 

\vspace{-10pt}
\paragraph{Evaluation metrics}
To evaluate our Aligner and Blender against the baselines, we use LPIPS \cite{zhang2018perceptual}, SSIM \cite{Wang2004ImageQA} and MS-SSIM \cite{1292216} to assess perceptual quality, and PSNR to measure reconstruction error. For Aligner, to compare source identity preservation, we compute cosine distance (CSIM) between embeddings of $I_A$ and $I_S$ from face recognition model \cite{deng2019arcface}. On cross-reenactment, we also utilize  Frechet Inception Distance (FID) \cite{NIPS2017_8a1d6947}. Additionally, for Blender we also measure reconstruction for background inpainting $\text{PSNR}_{\text{inpainting}}$ and head color transfer $\text{PSNR}_{\text{head}}$.

Additionally, in cross-reenactment scenario we conduct a user study to qualitatively compare preservation of source identity  (UAPP), transfer of target motion (UMTN) and overall quality (UQLT).

\subsection{Aligner evaluation}
\paragraph{Baselines} Our Aligner is compared against the open-source baselines at $512 \times 512$ and $256 \times 256$ resolutions. We note that the majority of competing models are trained on narrow face crops and hence do not reconstruct whole head and hair.

Few 2D reenactment models are available at $512 \times 512$ resolution. We compare against StyleHEAT \cite{yin2022styleheat}, based on StyleGAN \cite{karras2019style} inversion. Additionally, we train baseline from HeSer \cite{shu2022few} at $512 \times 512$ 

\FloatBarrier
\begin{table*}
\caption{Quantitative results on head reenactment at $512 \times 512$ and $256 \times 256$ resolution}
\label{sample-table}
\begin{center}
\begin{small}
\begin{tabular}{l|cccc|cc}
\toprule
\multirow{2}{*}{Method} & \multicolumn{4}{|c|}{Self-reenactment ($256 \times 256$)} & \multicolumn{2}{c}{Cross-reenactment ($256 \times 256$)} \\
 & CSIM $\uparrow$ & LPIPS $\downarrow$ & PSNR $\uparrow$ & SSIM $\uparrow$ & CSIM $\uparrow$ & FID $\downarrow$\\
\midrule
X2Face \cite{wiles2018x2face} & 0.789 & 0.201 & 19.66 & 0.715 & 0.636 & 41.95  \\
FOMM \cite{Siarohin_2019_NeurIPS} & \textbf{0.847} & 0.128 & 21.83 & 0.783 & 0.629 & 28.12  \\
PIRender \cite{ren2021pirenderer} & 0.827 & 0.173 & 19.46 & 0.704 & \textbf{0.665} & \textbf{19.95} \\
Bi-layer \cite{Zakharov20}  & 0.788 & 0.198 & 20.33 & 0.772 & 0.654 & 32.70\\
DaGAN \cite{hong2022depth} & 0.789 & 0.208 & 19.19 & 0.712 & 0.580 & 44.00 \\
DPE \cite{Pang_2023_CVPR} & \textbf{0.845} & 0.161 & 20.92 & 0.768 & 0.613 &  50.61 \\
LIA \cite{Wang2022LatentIA} & \textbf{0.842} & 0.128 & 22.11 & 0.786 & 0.659 & 24.19 \\
TPSMM \cite{zhao2022thin} & \textbf{0.843} &  0.124 & 21.86 & 0.804 & 0.647 & 22.06\\
\textbf{Ours} & 0.747 & \textbf{0.116} & \textbf{22.30} & \textbf{0.815} & 0.616 & 36.89     \\

\hline
\multirow{2}{*}{Method} & \multicolumn{4}{|c|}{Self-reenactment ($512 \times 512$)} & \multicolumn{2}{c}{Cross-reenactment ($512 \times 512$)} \\
 & CSIM $\uparrow$ & LPIPS $\downarrow$ & PSNR $\uparrow$ & SSIM $\uparrow$ & CSIM $\uparrow$ & FID $\downarrow$ \\
\midrule
HeSer \cite{shu2022few} & 0.780 &  \textbf{0.139} & \textbf{22.99} & \textbf{0.856} & 0.612 & 35.33   \\
StyleHEAT \cite{yin2022styleheat} & \textbf{0.789} &  0.574 & 8.791 & 0.582 & \textbf{0.673} & 75.856   \\
\textbf{Ours} & 0.754 & \textbf{0.142} & 22.04 & 0.848 & 0.628 & \textbf{29.57}  \\

\bottomrule
\end{tabular}
\end{small}
\end{center}
\vskip -0.1in
\label{tab:aligner}
\end{table*}
\FloatBarrier

resolution for full head synthesis.

At $256 \times 256$ resolution, we compare against the following warping-based approaches: First Order Motion Model (FOMM) \cite{Siarohin_2019_NeurIPS}, PIRenderer \cite{ren2021pirenderer}, Depth-Aware Generative Adversarial Network (DaGAN) \cite{hong2022depth}, Disentanglement of Pose and Expression (DPE) model \cite{Pang_2023_CVPR}, Latent Image Animator (LIA) \cite{Wang2022LatentIA} and Thin-Plate Spline Motion Model (TPSMM) \cite{zhao2022thin}. We also include methods based on latent face reconstruction with target motion: X2Face \cite{wiles2018x2face} and Fast Bi-layer Neural Synthesis \cite{zakharov2020fast}. All these methods generate only narrow face crops and not the whole head.

\begin{figure}
  \centering
  \begin{tabular}{*{5}{@{\hspace{0pt}}c}}
  Source & Target & \small{GHOST 2.0} & HeSer& \small{StyleHEAT}\\
  \includegraphics[width=0.2\columnwidth]{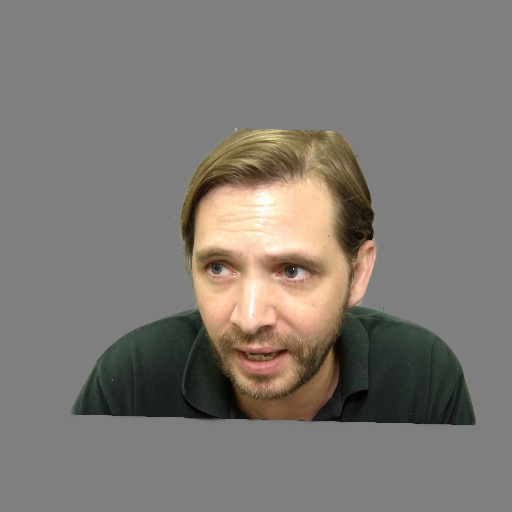}&
    \includegraphics[width=0.2\columnwidth]{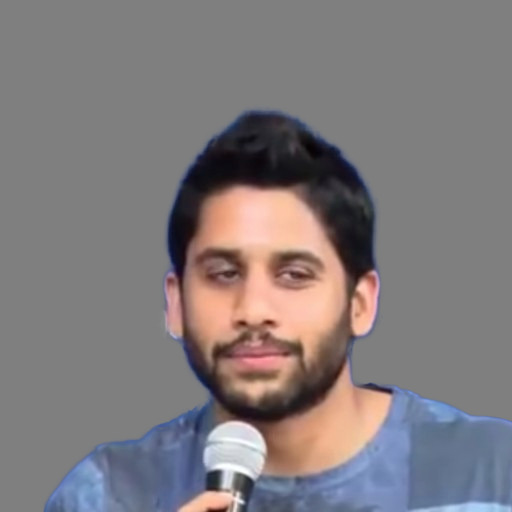}&
    \includegraphics[width=0.2\columnwidth]{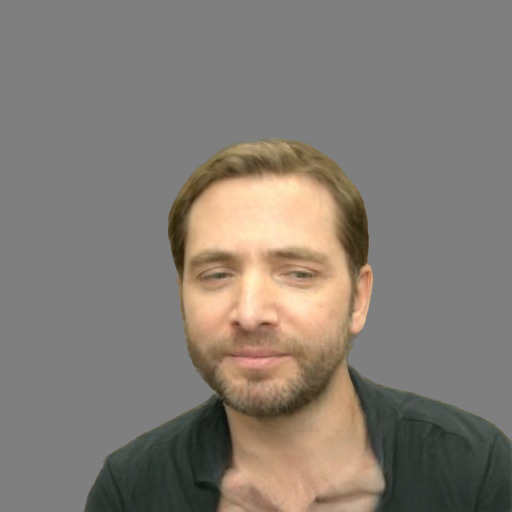}&
    \includegraphics[width=0.2\columnwidth]{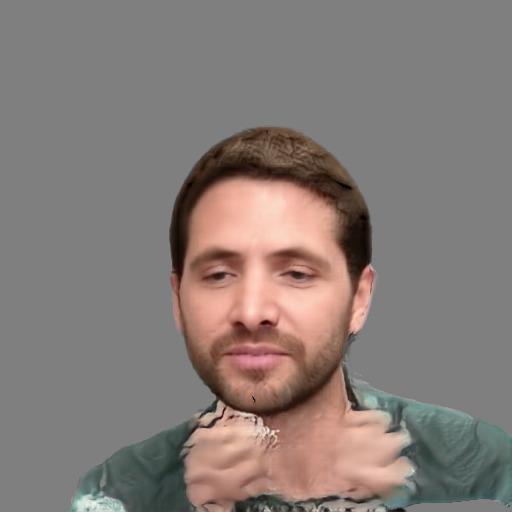}&
    \includegraphics[width=0.2\columnwidth]{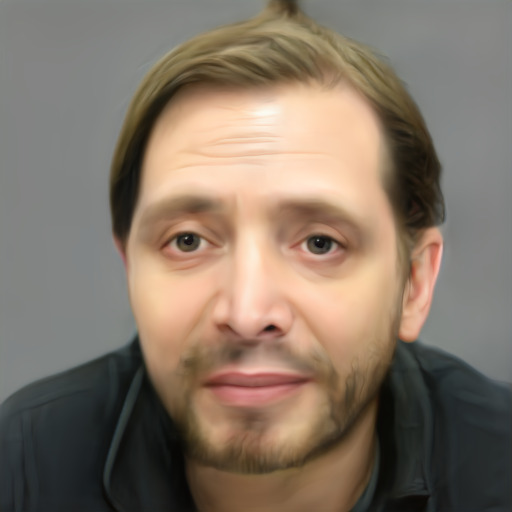}\\
  \includegraphics[width=0.2\columnwidth]{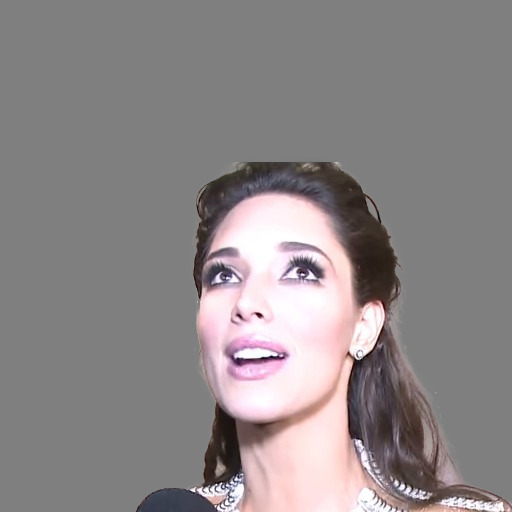}&
    \includegraphics[width=0.2\columnwidth]{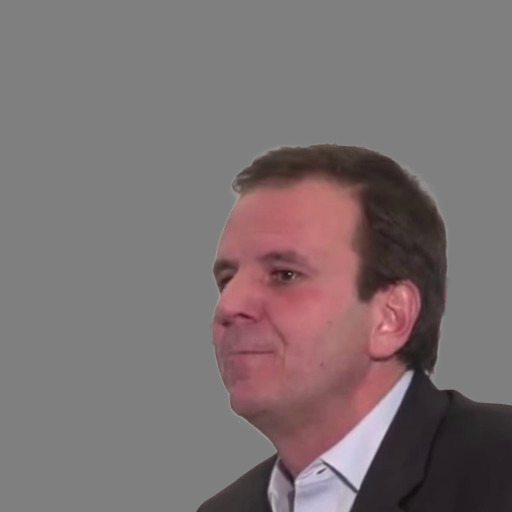}&
    \includegraphics[width=0.2\columnwidth]{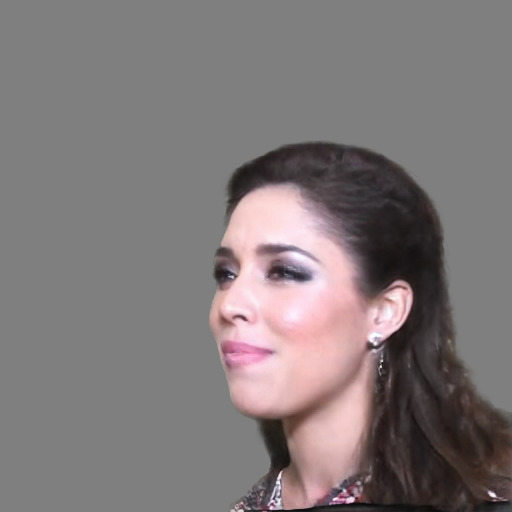}&
    \includegraphics[width=0.2\columnwidth]{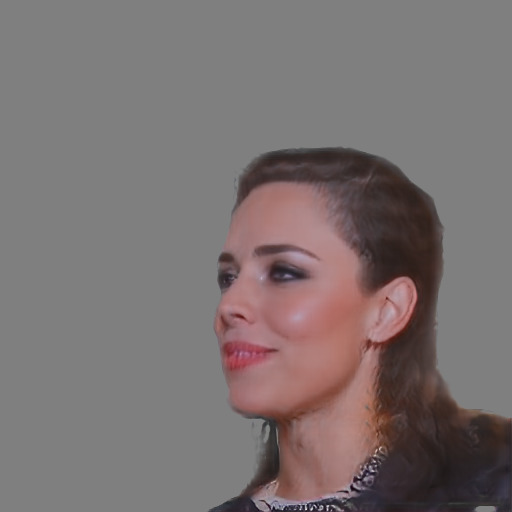}&
    \includegraphics[width=0.2\columnwidth]{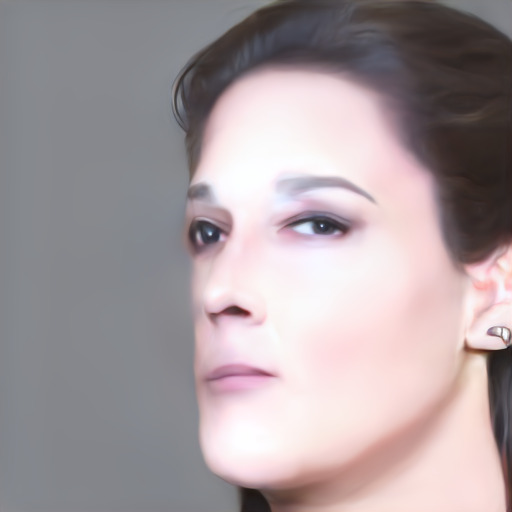}\\
  \includegraphics[width=0.2\columnwidth]{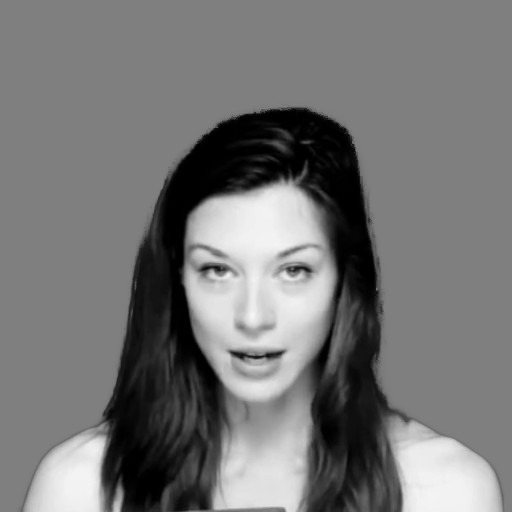}&
    \includegraphics[width=0.2\columnwidth]{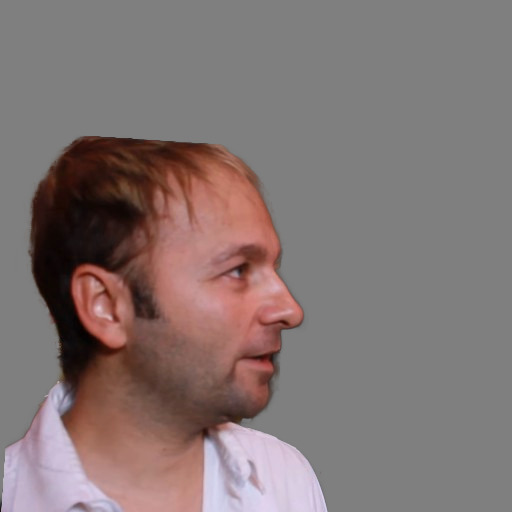}&
    \includegraphics[width=0.2\columnwidth]{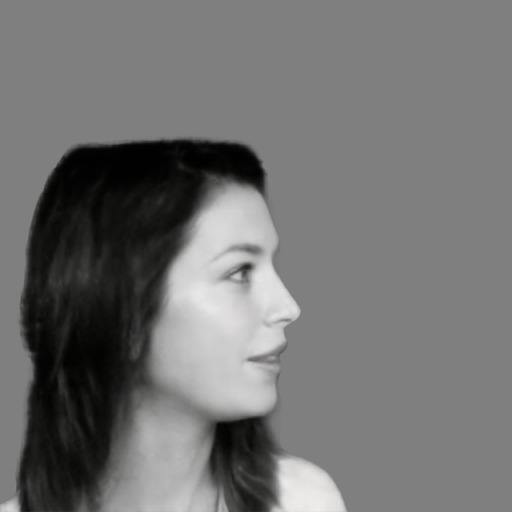}&
    \includegraphics[width=0.2\columnwidth]{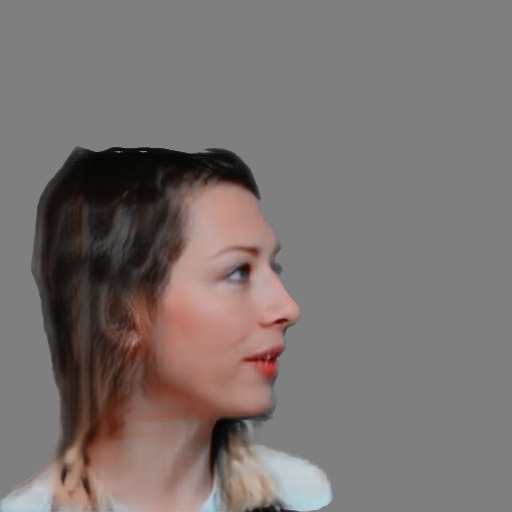}&
    \includegraphics[width=0.2\columnwidth]{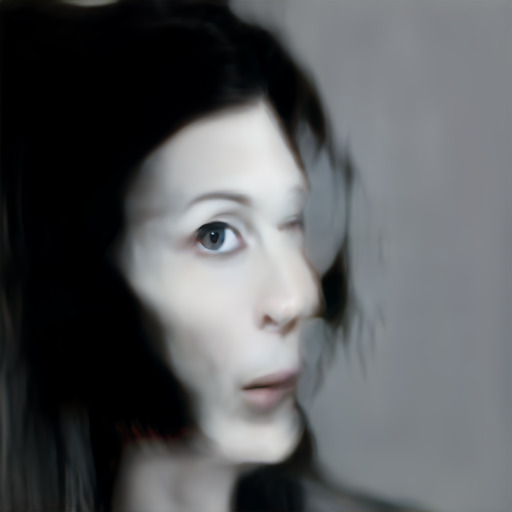}\\
    \includegraphics[width=0.2\columnwidth]{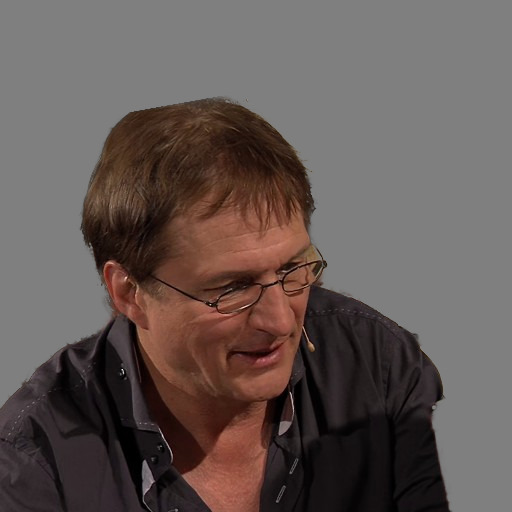}&
    \includegraphics[width=0.2\columnwidth]{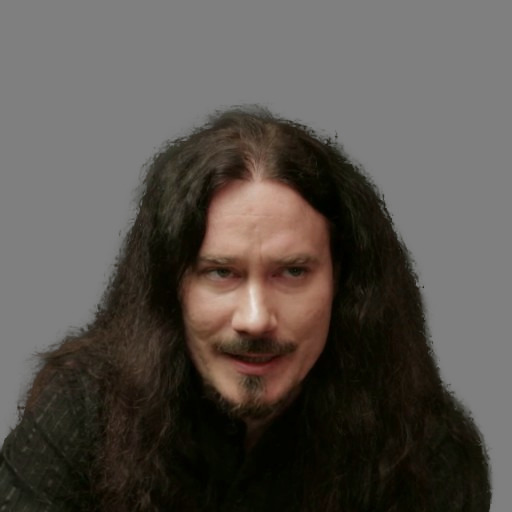}&
    \includegraphics[width=0.2\columnwidth]{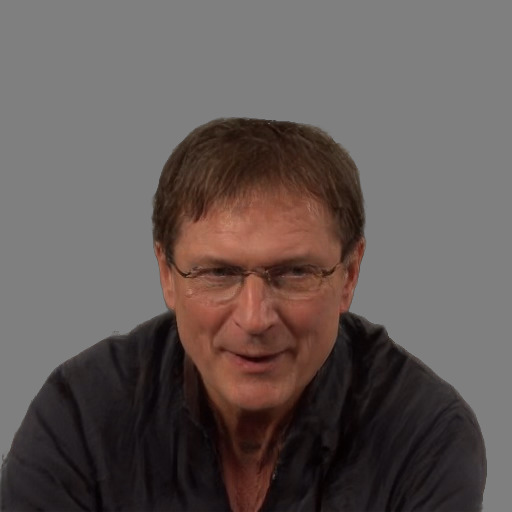}&
    \includegraphics[width=0.2\columnwidth]{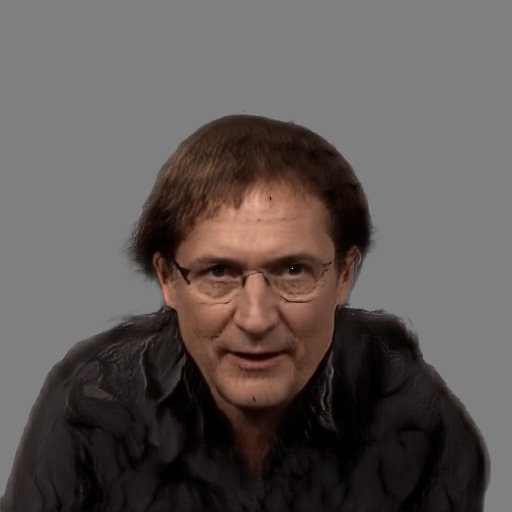}&
    \includegraphics[width=0.2\columnwidth]{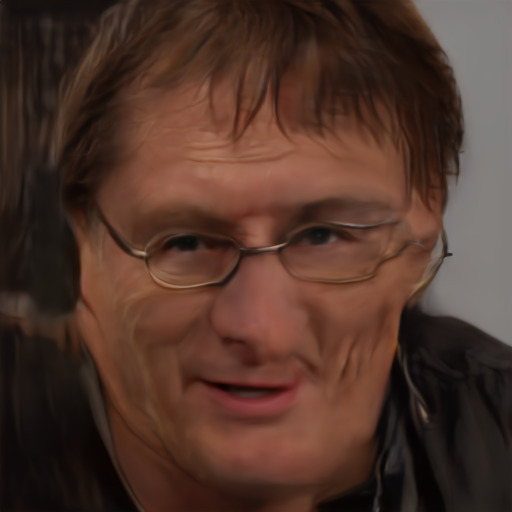}\\
    \includegraphics[width=0.2\columnwidth]{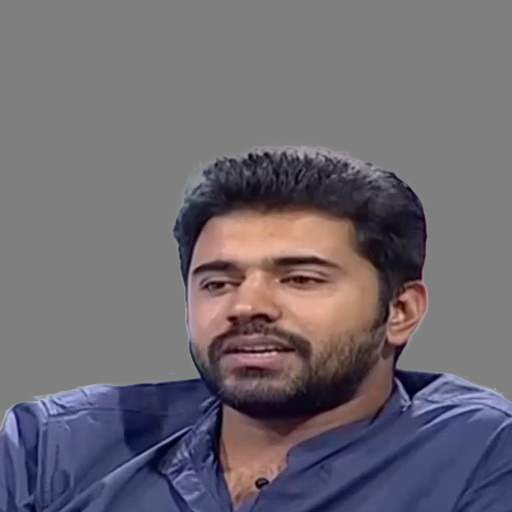}&
    \includegraphics[width=0.2\columnwidth]{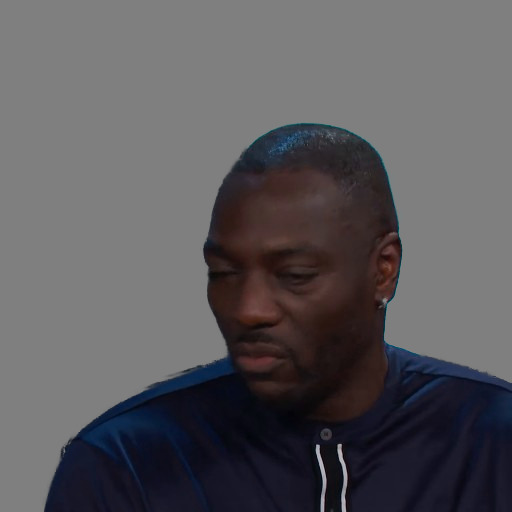}&
    \includegraphics[width=0.2\columnwidth]{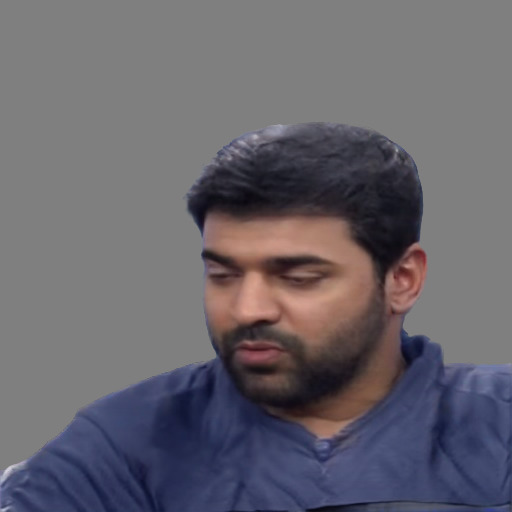}&
    \includegraphics[width=0.2\columnwidth]{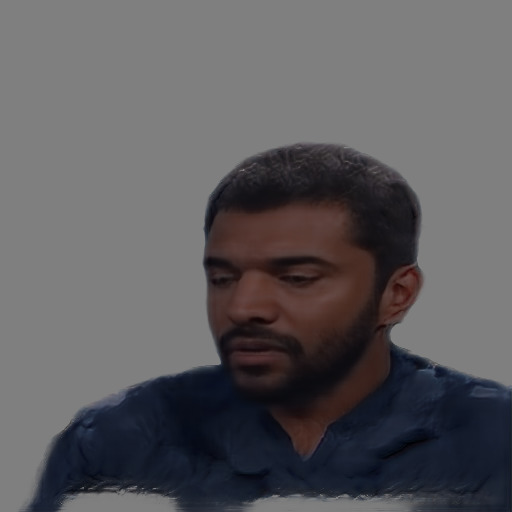}&
    \includegraphics[width=0.2\columnwidth]{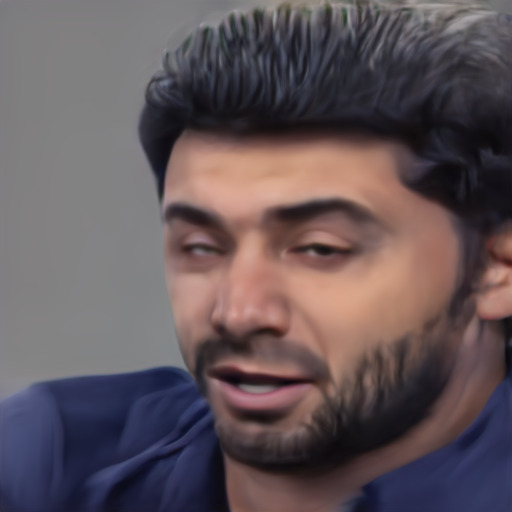}\\
    \includegraphics[width=0.2\columnwidth]{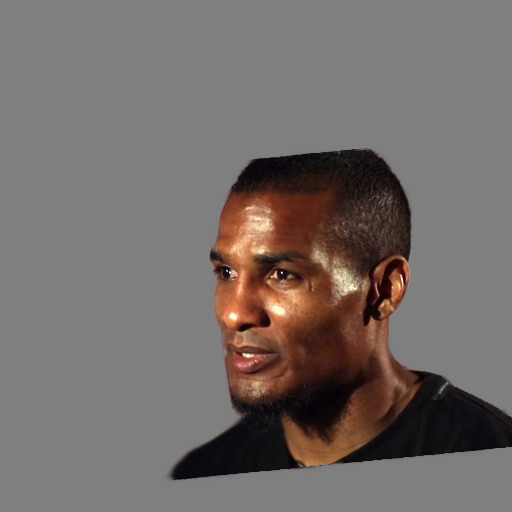}&
    \includegraphics[width=0.2\columnwidth]{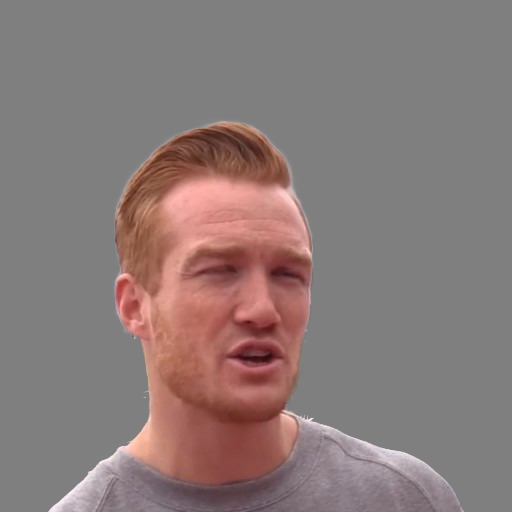}&
    \includegraphics[width=0.2\columnwidth]{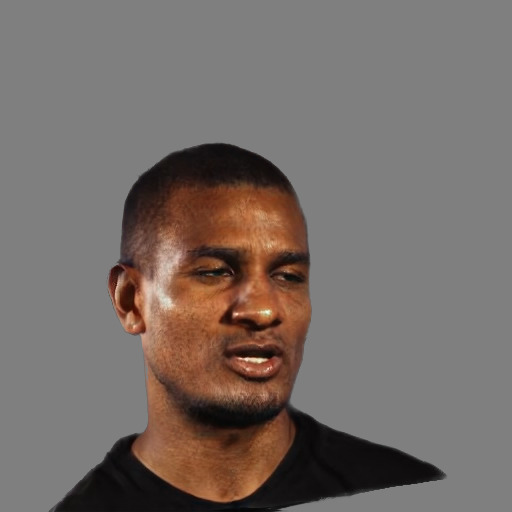}&
    \includegraphics[width=0.2\columnwidth]{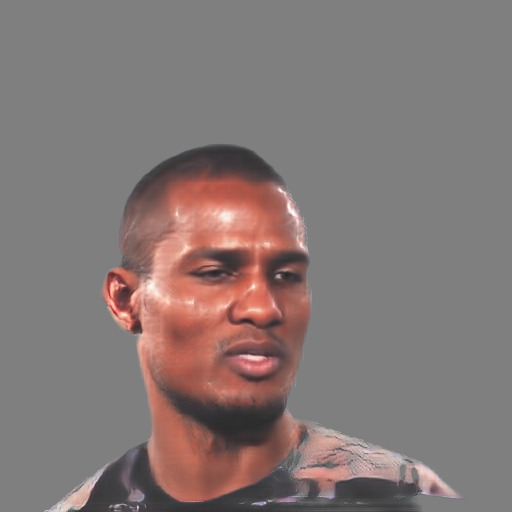}&
    \includegraphics[width=0.2\columnwidth]{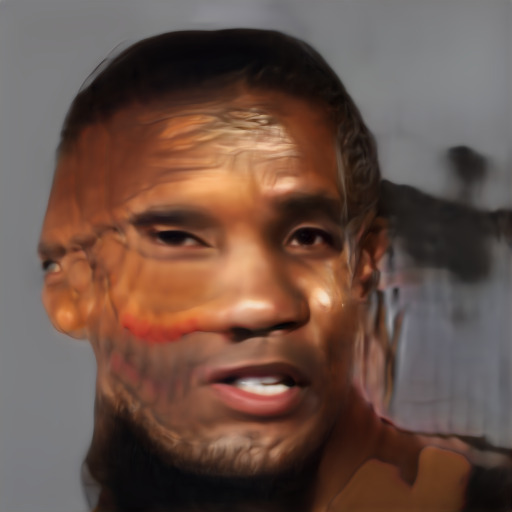}\\
  \end{tabular}
  \caption{Cross-reenactment results at $512 \times 512$ resolution}
  \label{fig:aligner_res}
\end{figure}

\paragraph{Results}
The results for self- and cross-reenactment scenarios at $512 \times 512$ and $256 \times 256$ resolutions are presented in table \ref{tab:aligner}. At $512 \times 512$ resolution, HeSer \cite{shu2022few} outperforms other methods by almost all metrics at self-reenactment. This is largely attributed to the leakage of target identity into final generation, which supplies additional information on the desired result. However, it is inferior to GHOST 2.0 on cross-reenactment. As can be seen from fig. \ref{fig:aligner_res}, GHOST 2.0 is significantly better at preserving source identity and skin color, while the HeSer produces a mixed identity of source and driver. Also, compared to StyleHEAT, our model is more robust to generation in extreme poses, although it may be slightly inferior in terms of identity preservation in frontal views.

At $256 \times 256$ resolution,  we outperform other methods by LPIPS, PSNR and SSIM in self-reenactment. This is in part explained by robustness of our method to generation in difficult poses. Warping-based methods perform well only in case of small displacements, resulting in severe face distortion and artifacts otherwise. However, they usually excel in identity preservation, as evidenced by CSIM both on self- and cross-reenactment. Please see supplement for visual comparison of the models.

\begin{table}
\caption{Side-by-side comparison at $512\times512$ resolution}
\label{sample-table}
\begin{center}
\begin{small}
\begin{tabular}{l|ccc} 
\toprule

Method & UAPP $\uparrow$ & UMTN $\uparrow$ & UQLT $\uparrow$ \\
\hline
HeSer \cite{shu2022few} & 0.04 & 0.15 & 0.04 \\
StyleHEAT \cite{yin2022styleheat} & 0.13 & 0.05 & 0.03 \\
\textbf{Ours} & \textbf{0.83} & \textbf{0.80} & \textbf{0.93} \\
\bottomrule
\end{tabular}
\end{small}
\end{center}
\label{tab:sbs512}
\end{table}
\begin{table}
\caption{Side-by-side comparison at $256\times256$ resolution}
\label{sample-table}
\begin{center}
\begin{small}
\begin{tabular}{l|ccc}
\toprule

Method & UAPP $\uparrow$ & UMTN $\uparrow$ & UQLT $\uparrow$ \\
\hline
Bi-layer \cite{Zakharov20} & \textbf{0.81} & \textbf{0.83} & \textbf{0.91} \\
DaGAN \cite{hong2022depth} & 0.11 & 0.13 & 0.07 \\
X2Face \cite{wiles2018x2face} & 0.08 & 0.04 & 0.02 \\
\hline
LIA \cite{Wang2022LatentIA} & \textbf{0.52} & \textbf{0.42} & \textbf{0.51} \\
TPSMM \cite{zhao2022thin} & 0.20 & 0.35 & 0.25 \\
PIRender \cite{ren2021pirenderer} & 0.28 & 0.23 & 0.24 \\
\hline
DPE \cite{Pang_2023_CVPR} & 0.27 & 0.13 & 0.12 \\
FOMM \cite{Siarohin_2019_NeurIPS} & 0.10 & 0.12 & 0.07 \\
\textbf{Ours} & \textbf{0.63} & \textbf{0.75} & \textbf{0.81} \\
\hline
Bi-layer \cite{Zakharov20} & 0.13 & 0.10 & 0.07 \\
LIA \cite{Wang2022LatentIA} & 0.41 & 0.29 & 0.29 \\
\textbf{Ours} & \textbf{0.46} & \textbf{0.61} & \textbf{0.64} \\
\bottomrule
\end{tabular}
\end{small}
\end{center}
\label{tab:sbs256}
\end{table}

\paragraph{Side-by-side}
We also conducted side-by-side comparison on cross-reenactment scenario between our Aligner and the competitors both at $256 \times 256$ and $512 \times 512$ resolutions. We asked the users to choose the model which performs best in terms of the following criteria: source identity preservation (UAPP), target movement transfer (UMTN) and overall generation quality (UQLT). We present the percentage of examples where each model is chosen in tables \ref{tab:sbs512} and \ref{tab:sbs256}. Since at $256 \times 256$ resolution we have a large number of models to compare against, we split them into three triplets first, compare within them, and then compare winners of these triplets. On average, each image from the test dataset is shown to 41 users.

Our method outperforms the competitors by a large margin in terms of motion preservation and generation quality at both resolutions. It is also significantly better than most of the methods in terms of identity preservation. Several methods, such as DPE \cite{Pang_2023_CVPR} or LIA \cite{Wang2022LatentIA}, fail at generation with large head rotations and produce blank 
outputs. Our method shows robustness to various head poses and expressions.

\paragraph{Ablation}
We conducted ablation study to justify our design of combined motion encoder $E_{mtn}$. We compared different strategies to disentangle pose and facial expression descriptors through special losses and architectural changes. The results are shown in table \ref{tab:aligner_abl}. In addition to FID and CSIM, we also use Average Keypoint Distance (AKD) to measure motion transfer. 

In HeSer, two separate MobileNetV2 \cite{sandler2018mobilenetv2} encoders are used to capture pose and expression. We also

\begin{figure}[H]
  \centering
  \begin{tabular}{*{3}{@{\hspace{0pt}}c}}
  Source & Target & \small{GHOST 2.0}\\
  \includegraphics[width=0.3\columnwidth]{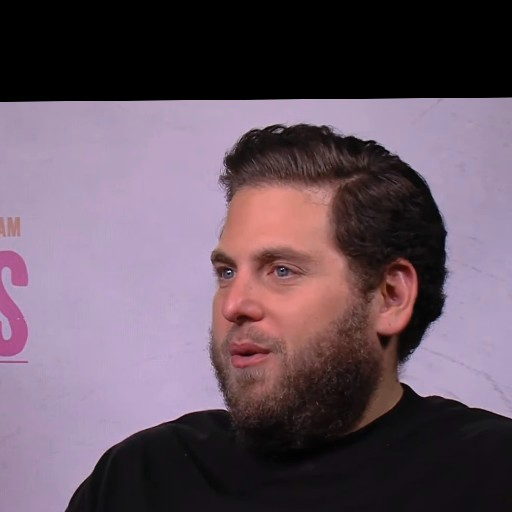}&
    \includegraphics[width=0.3\columnwidth]{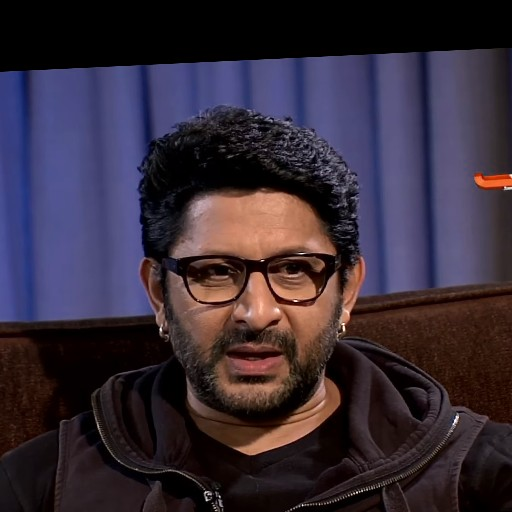}&
    \includegraphics[width=0.3\columnwidth]{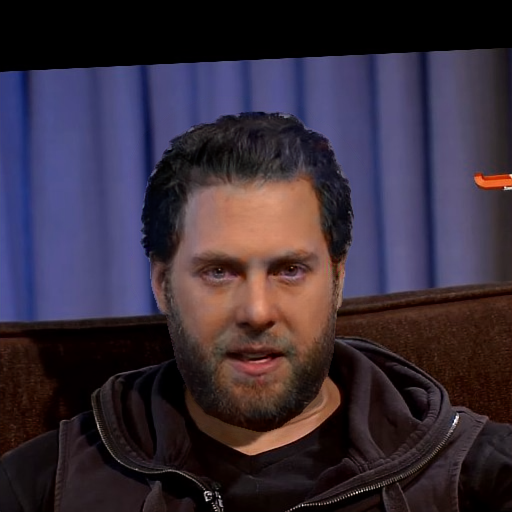}\\
  \includegraphics[width=0.3\columnwidth]{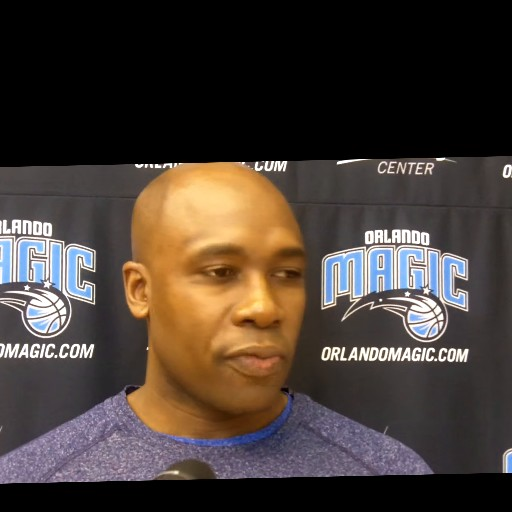}&
    \includegraphics[width=0.3\columnwidth]{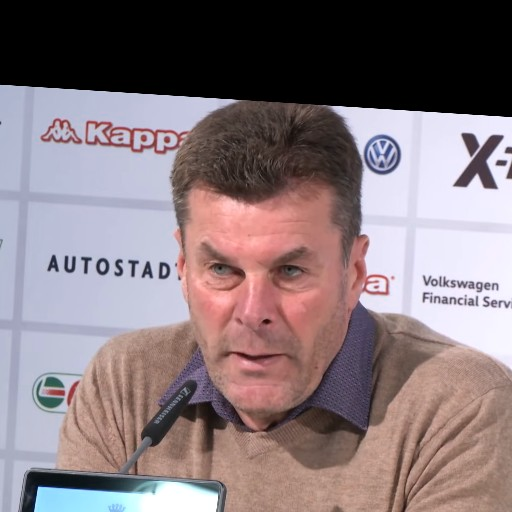}&
    \includegraphics[width=0.3\columnwidth]{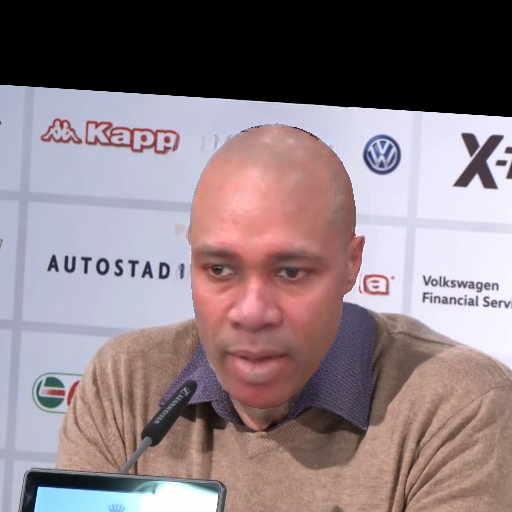}\\
    \includegraphics[width=0.3\columnwidth]{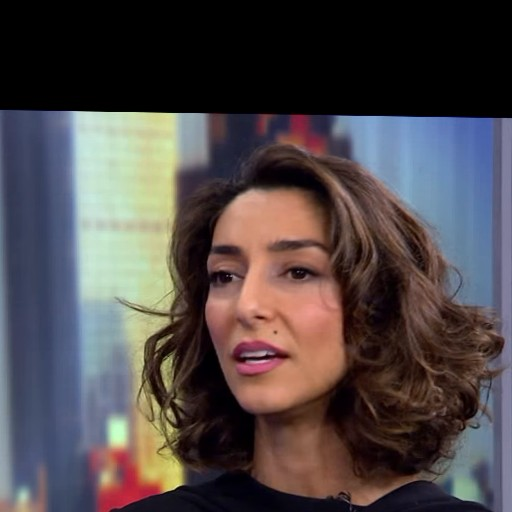}&
    \includegraphics[width=0.3\columnwidth]{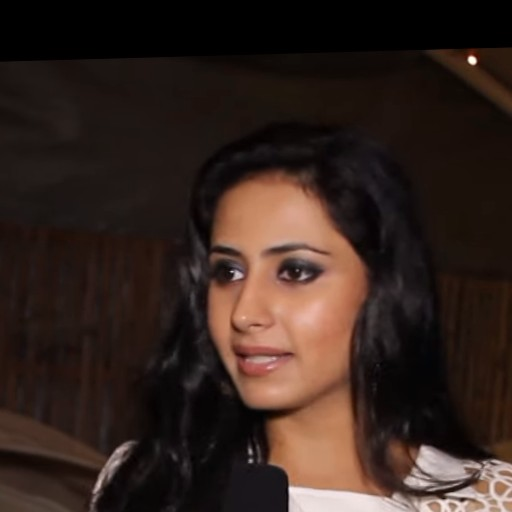}&
    \includegraphics[width=0.3\columnwidth]{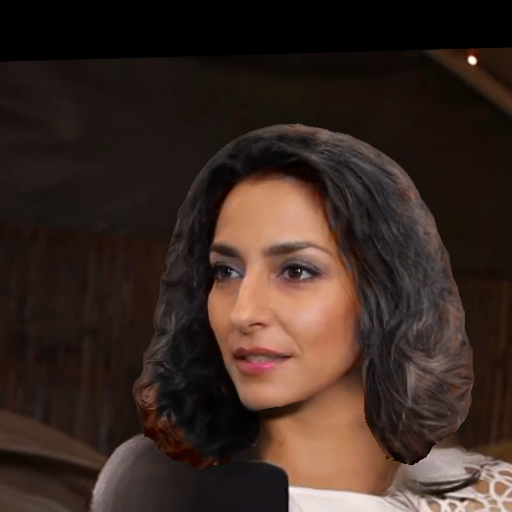}\\
    \includegraphics[width=0.3\columnwidth]{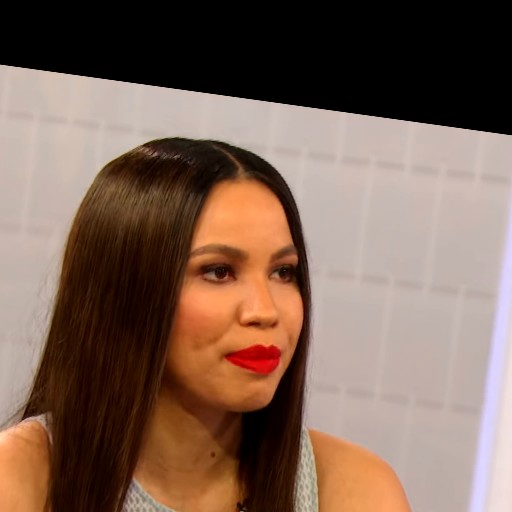}&
    \includegraphics[width=0.3\columnwidth]{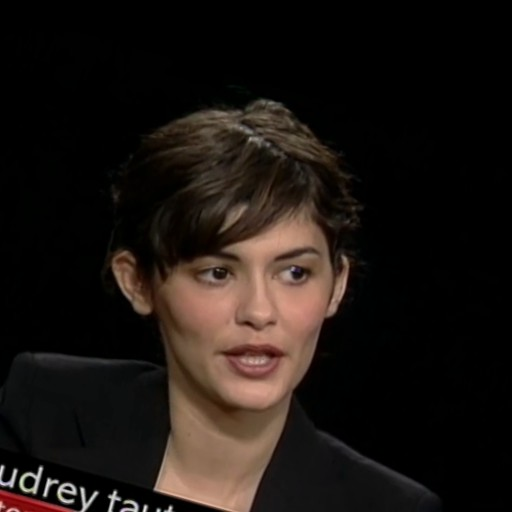}&
    \includegraphics[width=0.3\columnwidth]{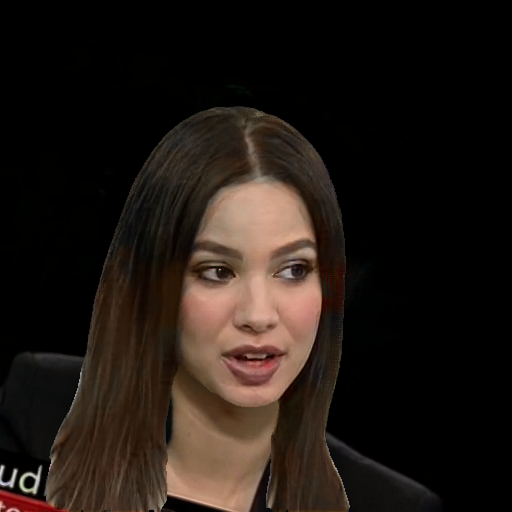}\\
    \includegraphics[width=0.3\columnwidth]{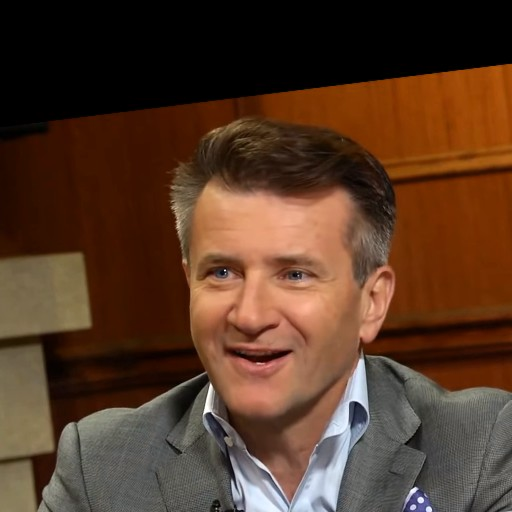}&
    \includegraphics[width=0.3\columnwidth]{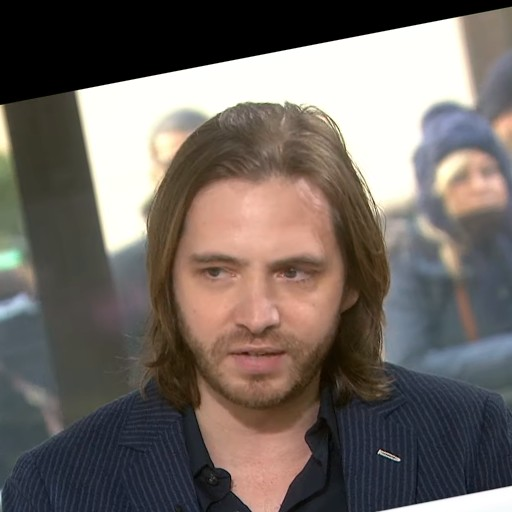}&
    \includegraphics[width=0.3\columnwidth]{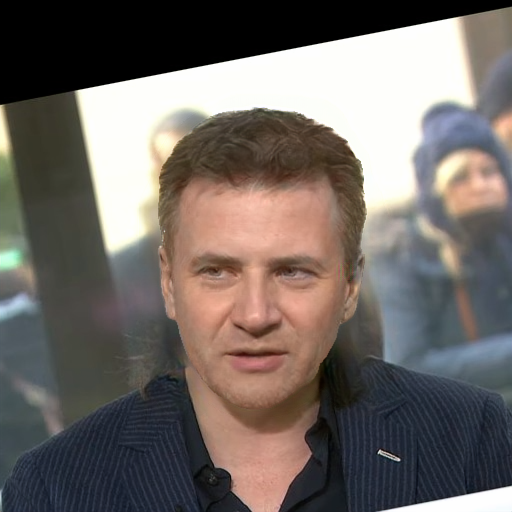}\\
    \includegraphics[width=0.3\columnwidth]{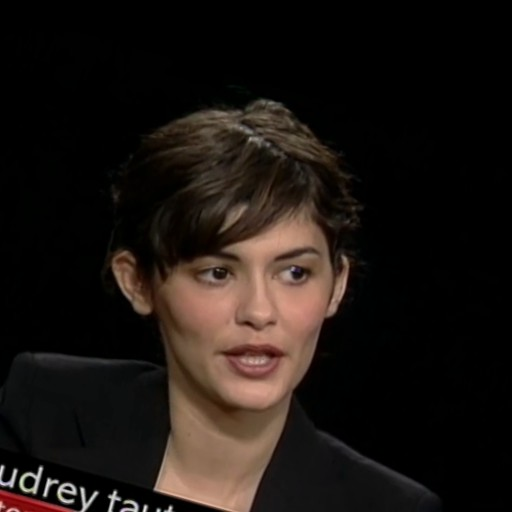}&
    \includegraphics[width=0.3\columnwidth]{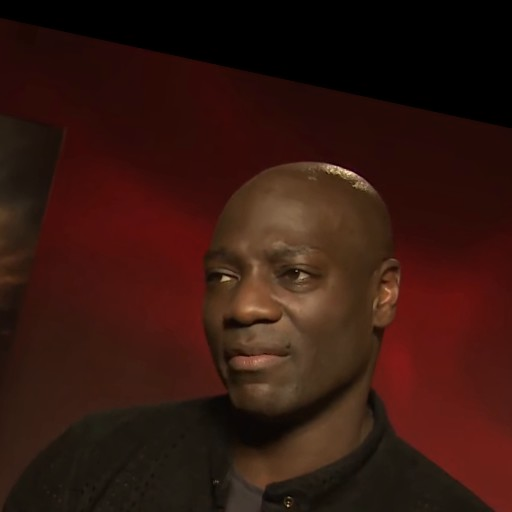}&
    \includegraphics[width=0.3\columnwidth]{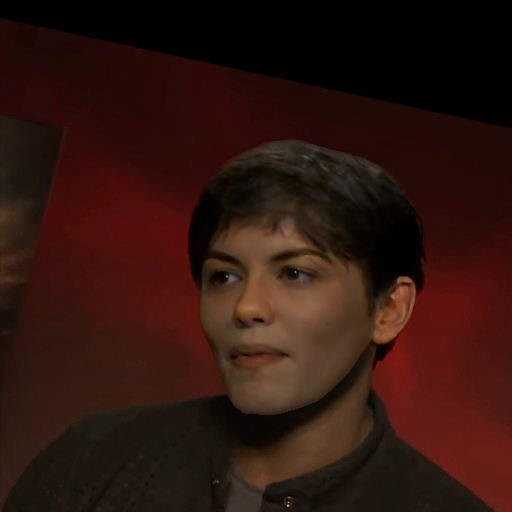}\\
    \includegraphics[width=0.3\columnwidth]{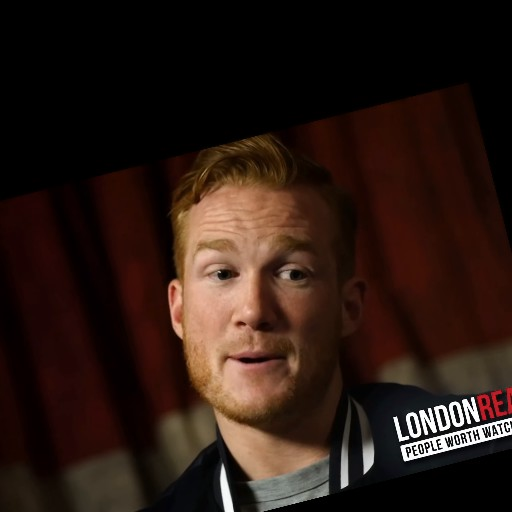}&
    \includegraphics[width=0.3\columnwidth]{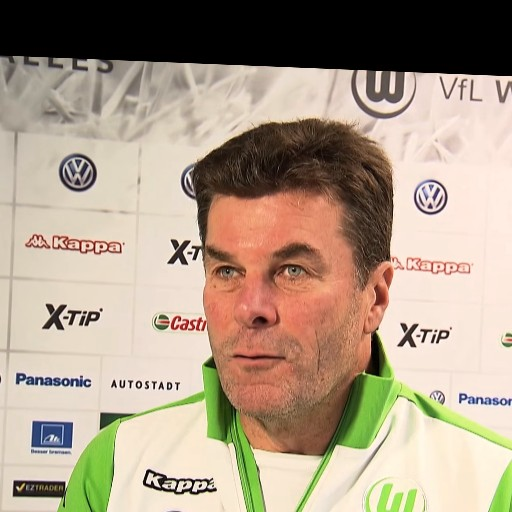}&
    \includegraphics[width=0.3\columnwidth]{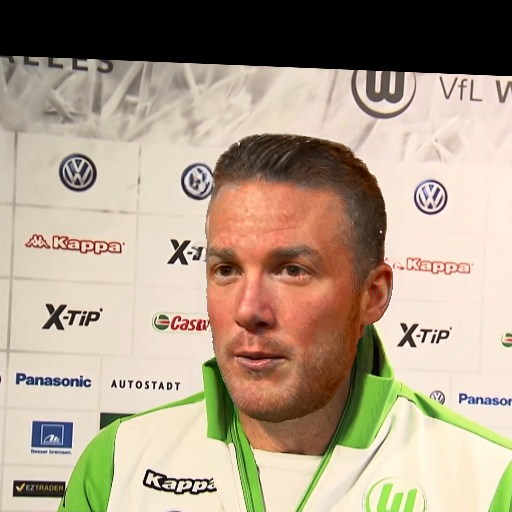}\\
  \end{tabular}
  \caption{GHOST 2.0 results on the task of head swap. Our method achieves natural blending of the reenacted head into target background, corresponding in skin color, lighting and contrast to the rest of the image.}
  \label{fig:blender_res}
\end{figure}

 conducted experiment with decreasing their number of channels in half to force them to learn only relevant information. In other experiments, we replaced them with corresponding pretrained encoders from DECA \cite{DECA:Siggraph2021}, and  
\FloatBarrier
\begin{table*}
\caption{Quantitative results on head blending}
\label{blender-table}
\begin{center}
\begin{small}
\begin{tabular}{l|cccccc}
\toprule
\multirow{2}{*}{Method} & \multicolumn{6}{c}{Plain data} \\
  & LPIPS $\downarrow$ & SSIM $\uparrow$ & MSSIM $\uparrow$ & PSNR $\uparrow$ & $\text{PSNR}_{\text{inpainting}}$ $\uparrow$ & $\text{PSNR}_{\text{head}}$ $\uparrow$ \\
\midrule
Baseline & 0.034 & 0.972 & 0.987 & 33.86 & 35.13 & 42.13 \\
REFace & 0.24 & 0.89 & 0.78 & 23.45 & 36.72 & 29.29 \\
InstantID + SDXL inpainting & 0.10 & 0.94 & 0.91 & 29.49 & \textbf{38.54} & 30.38 \\
External inpainting & \textbf{0.018} & \textbf{0.984} & \textbf{0.992} & \textbf{36.65} & \textbf{37.98} & \textbf{45.83}\\

\hline
\multirow{2}{*}{Method} & \multicolumn{6}{c}{Mask augmentations and source color jitter}  \\
 & LPIPS $\downarrow$ & SSIM $\uparrow$ & MSSIM $\uparrow$ & PSNR $\uparrow$ & $\text{PSNR}_{\text{inpainting}}$ $\uparrow$ & $\text{PSNR}_{\text{head}}$ $\uparrow$ \\
\midrule
Baseline & 0.093 &  0.926 & 0.945 & 27.76 & 28.15 & 41.25  \\
External inpainting & \textbf{0.060} &  \textbf{0.947} & \textbf{0.951} & \textbf{28.21} & \textbf{28.45} & \textbf{44.13} \\

\bottomrule
\end{tabular}
\end{small}
\end{center}
\vskip -0.1in
\end{table*}
\FloatBarrier

also tried to disentangle representations with cyclic loss from DPE \cite{Pang_2023_CVPR}. Our design with combined encoder $E_{mtn}$ achieves the best quality by all metrics. Further details on other ablations and calculation of AKD are given in supplementary.

\begin{table}[H]
\caption{Results on cross-reenactment for different disentanglement strategies between target motion and identity}
\label{sample-table}
\begin{center}
\begin{small}
\begin{tabular}{p{3.6cm}|p{0.8cm}p{1cm}p{0.9cm}}
\toprule

Method & FID $\downarrow$ & CSIM $\uparrow$ & AKD $\downarrow$ \\
\hline
HeSer \cite{shu2022few} & 28.63 & 0.596 & \textbf{0.0095} \\
Smaller encoders & 28.38 & \textbf{0.622} & 0.0108 \\
DECA encoders \cite{DECA:Siggraph2021} & 35.41 & \textbf{0.627} & 0.0155 \\
DPE loss \cite{Pang_2023_CVPR} & 31.70 & 0.603 & 0.0103 \\
Combined $E_{mtn}$ & \textbf{26.83} & \textbf{0.622} & \textbf{0.0098} \\
\bottomrule
\end{tabular}
\end{small}
\end{center}
\label{tab:aligner_abl}
\vskip -0.1in
\end{table}

\subsection{Blender evaluation}
To evaluate our Blender and justify external inpainting usage in it, we trained baseline HeSer Blender on the same data. Additionally, we compare with diffusion-based head swap approaches, such as REFace \cite{baliah2025realistic} and InstantID \cite{wang2024instantid}. For the latter solution, we first use InstantID \cite{wang2024instantid} to generate source head with target pose directed by facial keypoints. Then the resulting head is blended into target background using SDXL \cite{podell2023sdxl} inpainting model. It should be noted that both solutions frequently fail to preserve identity and target skin color. As can be seen from table \ref{blender-table}, our version with external inpainting outperforms baseline and competitors on almost every metric. However, for background inpainting, SDXL shows slightly better results than LaMa, so further investigation into better inpainting models may constitute an area for future research.  Furthermore, for external inpainting and baseline, we present results on training with additional data augmentations. Our solution outperforms baseline in this scenario as well and also doesn't fail in hard cases when source and target differ significantly (fig. \ref{fig:blender_res}).

In addition,  we present inference results of our model on real-life and outdoor photos in supplementary \ref{real}. Although our model successfully copes with most cases, sometimes Kandinsky model hallucinates during post-processing and may produce additional attributes such as collars when inpainting excess hair. Also, hair for reenacted source head is generated only inside the crop region. In case target person has shorter hair, it is likely that in the resulting image source hair would look unnaturally cropped from below.




\section{Conclusion}
We have presented a two-stage method for realistic head swapping for in-the-wild images. We improve Aligner architecture by merging pose and expression encoders into single motion encoder, which remedies the problem of driver identity leakage. Our head reenactment model outperforms other methods by both qualitative and quantitative metrics and is more robust to large pose variations.
We also introduce additional refinements during blending to improve quality of head transfer and inpainting, which allows to obtain superior results compared to baseline solution.

The limitations of our model are the following. In several cases, our method does not reproduce fine details of source appearance. This can be tackled by using stronger appearance encoders in Aligner. Concerning Blender, some face parts may not be evenly colored if the area of corresponding color reference is small. Additionally, blending stage could be improved by mitigating the generation of additional clothing attributes (e.g. collars) and better hair processing.
Complex lighting conditions can also produce inferior results during blending.

\section{Impact Statement}
This paper enhances approach to head swapping via better head reenactment and inpainting modules.  
While such models find applications in commercial scenarios, they are also known to be used for fraudulent activities. However, we suppose that results of this work can be used to fight such misuse by aiding research on more robust deepfake detection systems. 

\section{Acknowledgments}
We thank Nikolay Gerasimenko, Anna Averchenkova and ABT data labelling team for help with side-by-side comparison. We also thank Viacheslav Vasilev for suggestions and comments regarding text. Last but not least, we thank Alexander Kapitanov and his team from SberDevices for training of segmentation model. 

\bibliography{main}
\bibliographystyle{icml2025}

\newpage
\appendix
\clearpage
\section{Details on losses}
For training Aligner, we use the following combination of losses:
\begin{equation}
\begin{split}
    L_{aligner} = \lambda_{adv}\mathcal{L}_{adv} + \lambda_{FM}\mathcal{L}_{FM} + 
    \lambda_{L1}\mathcal{L}_{L1} \\+
    \lambda_{perc}^{VGG}\mathcal{L}_{perc}^{VGG}  +
    \lambda_{perc}^{ID}\mathcal{L}_{perc}^{ID} +
    \lambda_{cos}^{ID}\mathcal{L}_{cos}^{ID} +
    \lambda_{dice}\mathcal{L}_{dice} +\\
    \lambda_{emo}\mathcal{L}_{emo} +
    \lambda_{kpt}\mathcal{L}_{kpt} +
    \lambda_{gaze}\mathcal{L}_{gaze}
\end{split}
\end{equation}
We set the following weights for the losses: $\lambda_{adv}=0.1$, $\lambda_{FM}=10$, $\lambda_{L1}=30$, $\lambda_{perc}^{VGG}=0.01$, $\lambda_{perc}^{ID}=2\times 10^{-3}$,  $\lambda_{cos} ^{ID}=0.01$, $\lambda_{dice}=1$,  $\lambda_{emo}=1$, $\lambda_{kpt}=30$ and $\lambda_{gaze}=0.5$
For training Blender, we use the following losses:
\begin{equation}
\begin{split}
    L_{blender} = \lambda_{adv}\mathcal{L}_{adv} + 
    \lambda_{L1}\mathcal{L}_{L1} +
    \lambda_{perc}^{VGG}\mathcal{L}_{perc}^{VGG} \\ +
    \lambda_{c}\mathcal{L}_{c} +
    \lambda_{c}\mathcal{L}_{c'} +
    \lambda_{reg}\mathcal{L}_{reg}
\end{split}
\end{equation}
The weighs are set as $\lambda_{adv}=1$, $\lambda_{L1}=1$, $\lambda_{perc}^{VGG}=0.01$, $\lambda_{c}=1$ and $\lambda_{reg}=1$.

\section{Training details}
We trained Aligner for 1 000 000 iterations with batch size of 20 on 512x512 resolution, and for 800 000 iterations with batch size of 32 on 256x256 resolution. On 8 NVIDIA A 100 GPUs, it takes 27 and 9 days respectively. We use Adam optimizer \cite{kingma2014adam} with generator learning rate $1\times10^{-4}$ and discriminator learning rate $4\times10^{-4}$, with gradient clipping threshold of 10. We also apply spectral normalization \cite{miyato2018spectral} to stabilize training. 

We trained Blender for 50000 iterations with batch size of 20 on 512x512 resolution. On 4 NVIDIA A 100 GPUs it takes 2 days. We use Adam optimizer and same optimizer options as in Aligner.

\section{Detailed architecture}
\paragraph{Aligner} We use ResNeXt-50 \cite{xie2017aggregated} as our portrait encoder $E_{por}$, IResNet-50 \cite{duta2021improved} pretrained with Arcface \cite{deng2019arcface} loss as identity encoder $E_{id}$ and MobileNetV2 \cite{sandler2018mobilenetv2} as motion encoder $E_{motion}$. The dimensions of embeddings produced by these encoders are 512, 512 and 256, respectively. 

These embeddings are concatenated and processed with 2-layer MLP with ReLU activation and spectral normalization. The intermediate dimension is maintained the same as the input one. The resulting vector is supplied to AdaIn \cite{huang2017adain} layers to condition the generator, which is borrowed from \cite{Burkov_2020_CVPR}. We add one additional upsampling residual block to the original generator to increase output resolution from 256 to 512. Discriminator is also borrowed from \cite{Burkov_2020_CVPR} in its default version.

\section{Further ablations on Aligner}
We calculate Average Keypoint Distance (AKD) using keypoints from DECA \cite{DECA:Siggraph2021} model. Given a triplet of source $I_S$, target $I_T$ and generated $I_A$ images, we compute absolute distance between pair of normalized keypoints, which are based on shape blensdshapes from $I_S$ and pose and expression blendshapes from $I_T$ and $I_A$. In this way, we assess motion transfer, while keeping source appearance invariant.

We also show the effect of adding keypoint $\mathcal{L}_{kpt}$ and emotion $\mathcal{L}_{emo}$ losses when training motion encoder $E_{mtn}$. As can be seen in table \ref{tab:abl_loss}, they improve image quality, identity preservation and pose transfer in cross-reenactment scenario, and generally lead to better disentanglement between target motion and identity. On self-reenactment, metrics also generally improve with addition of these losses.

Finally, we also justify our choice to include gaze loss $\mathcal{L}_{gaze}$ only at the end of training after 1000 epochs in table \ref{tab:aligner_gaze}. We compare it to the experiment when we include $\mathcal{L}_{gaze}$ only after 10 training epochs, when the model is capable to generate eyes with enough details. Early addition of this loss results in a significant deterioration of source identity preservation and in a noticeable fall in general quality of images.

\begin{table}
\caption{Ablation of addition of keypoint $\mathcal{L}_{kpt}$ and emotion $\mathcal{L}_{emo}$ losses}
\begin{center}
\begin{small}
\begin{tabular}{l|ccccc}
\toprule


\multirow{2}{*}{Experiment} & \multicolumn{5}{c}{Cross-reenactment} \\
 & CSIM $\uparrow$ & FID $\downarrow$ & AKD $\downarrow$ & \\
\midrule
with losses & \textbf{0.621} &\textbf{26.83} & \textbf{0.0098} \\
w/o losses & 0.607 & 28.60 & 0.0107 \\
\hline
\multirow{2}{*}{Experiment} & \multicolumn{5}{c}{Self-reenactment} \\
 & CSIM $\uparrow$ & LPIPS $\downarrow$ & PSNR $\uparrow$ & SSIM $\uparrow$ & AKD $\downarrow$\\
\midrule
with losses & \textbf{0.745} & \textbf{0.150} & 21.87 & \textbf{0.845} & \textbf{0.0074}\\
w/o losses & 0.724 & 0.154 & \textbf{21.97} & 0.841 & 0.0079 \\

\bottomrule
\end{tabular}
\end{small}
\end{center}
\label{tab:abl_loss}
\vskip -0.1in
\end{table}

\begin{table}
\caption{Ablation on the start epoch for gaze loss $\mathcal{L}_{gaze}$}
\label{sample-table}
\begin{center}
\begin{small}
\begin{tabular}{l|ccccc}
\toprule

\multirow{2}{*}{Experiment} & \multicolumn{5}{c}{Cross-reenactment} \\
 & CSIM $\uparrow$ & FID $\downarrow$ & AKD $\downarrow$ & \\
\midrule
    Epoch 1000 & \textbf{0.621} & \textbf{26.83} & 0.0098 \\
    Epoch 10 & 0.538 & 37.22 & \textbf{0.0091} \\
\hline
\multirow{2}{*}{Experiment} & \multicolumn{5}{c}{Self-reenactment} \\
 & CSIM $\uparrow$ & LPIPS $\downarrow$ & PSNR $\uparrow$ & SSIM $\uparrow$ & AKD $\downarrow$\\
\midrule
    Epoch 1000 & \textbf{0.745} & \textbf{0.150} & \textbf{21.87} & \textbf{0.845} & \textbf{0.0074}\\
Epoch 10 & 0.699 & 0.166 & 21.47 & 0.837 & 0.0082 \\

\bottomrule
\end{tabular}
\end{small}
\end{center}
\label{tab:aligner_gaze}
\vskip -0.1in
\end{table}


\clearpage
\begin{figure*}[t!]
  \centering
  \begin{tabular}{*{5}{@{\hspace{0pt}}c}}
  Source & Target & \small{GHOST 2.0} & HeSer \cite{shu2022few}& \tiny{StyleHEAT \cite{yin2022styleheat}}\\
  \includegraphics[width=0.2\linewidth]{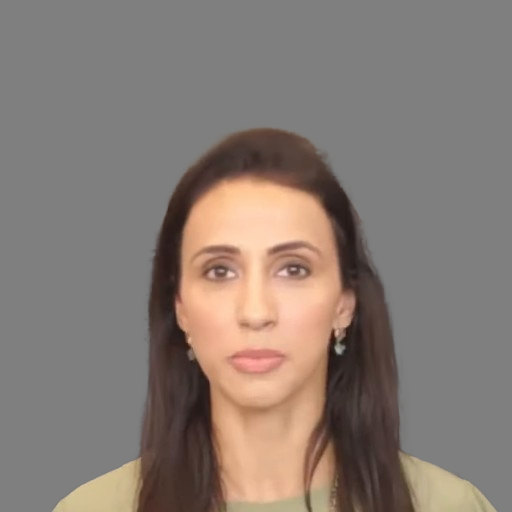}&
    \includegraphics[width=0.2\linewidth]{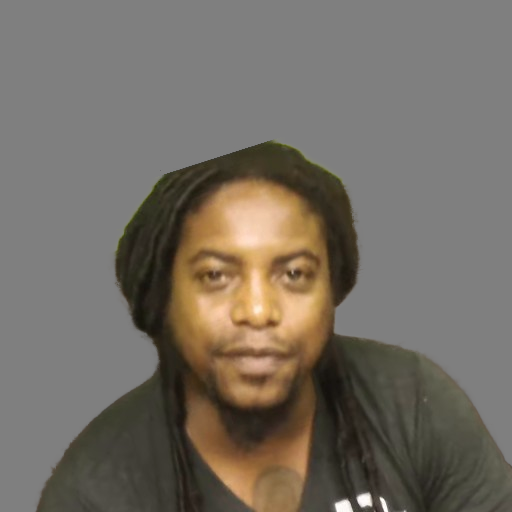}&
    \includegraphics[width=0.2\linewidth]{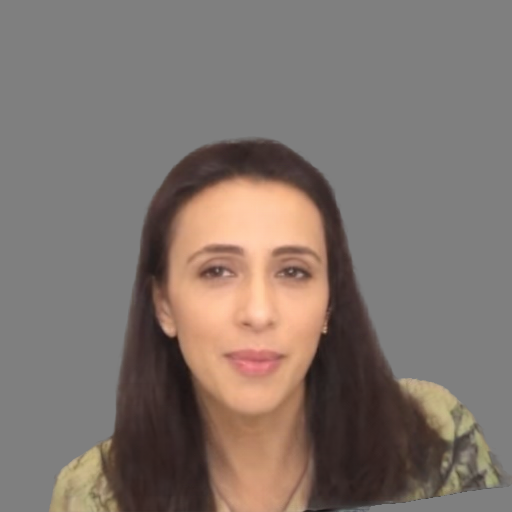}&
    \includegraphics[width=0.2\linewidth]{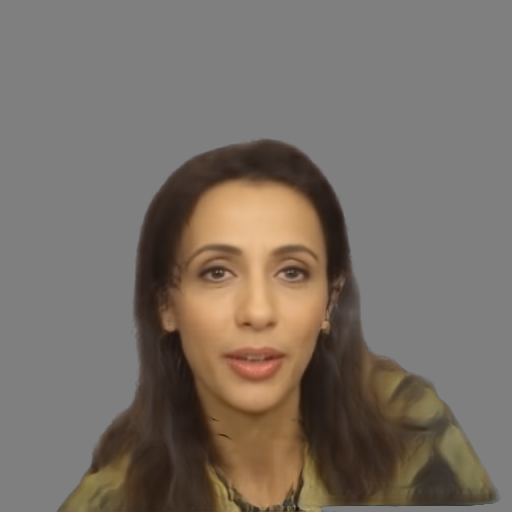}&
    \includegraphics[width=0.2\linewidth]{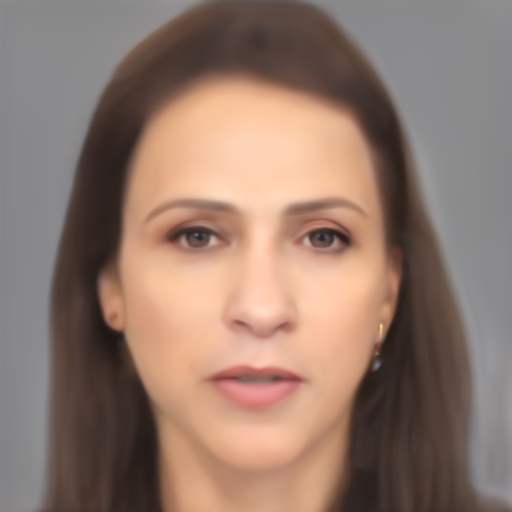}\\
    \includegraphics[width=0.2\linewidth]{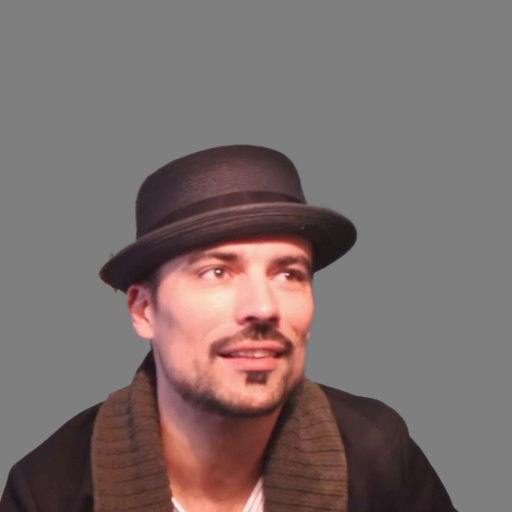}&
    \includegraphics[width=0.2\linewidth]{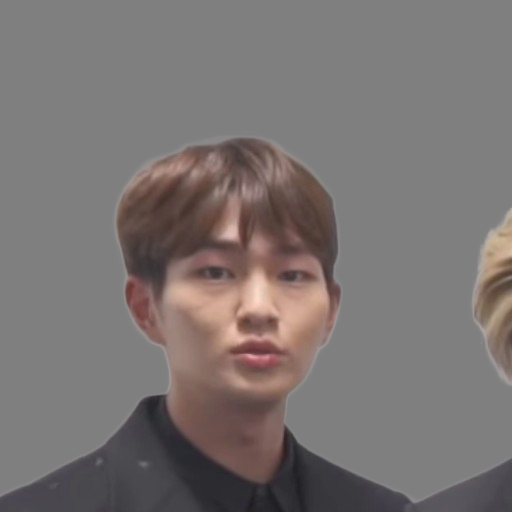}&
    \includegraphics[width=0.2\linewidth]{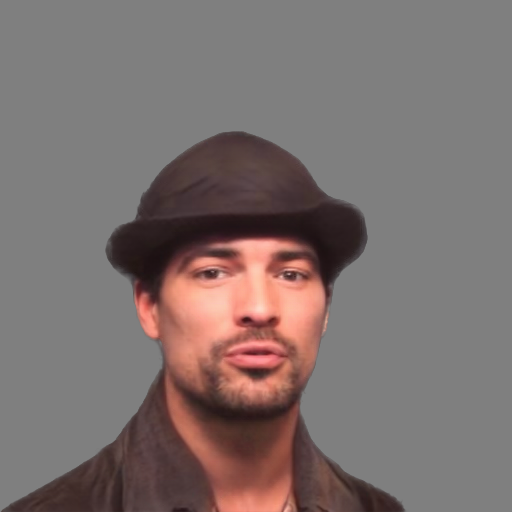}&
    \includegraphics[width=0.2\linewidth]{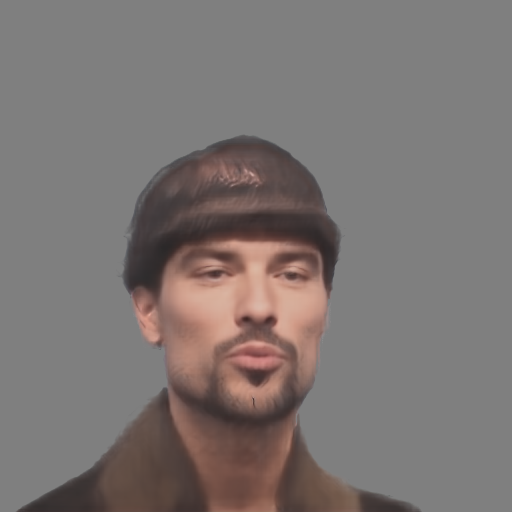}&
    \includegraphics[width=0.2\linewidth]{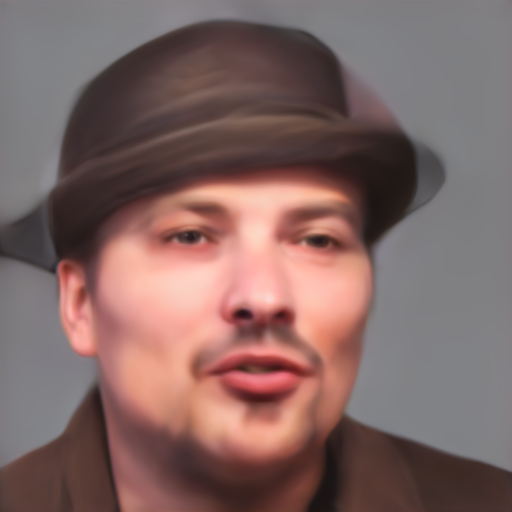}\\
    \includegraphics[width=0.2\linewidth]{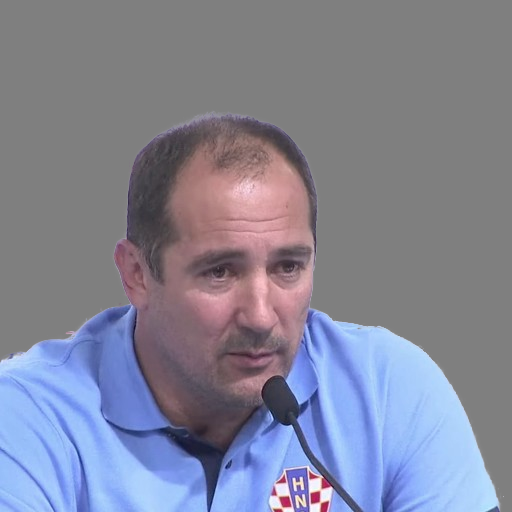}&
    \includegraphics[width=0.2\linewidth]{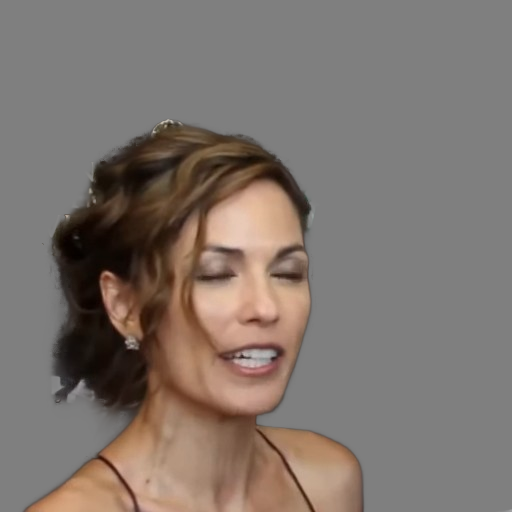}&
    \includegraphics[width=0.2\linewidth]{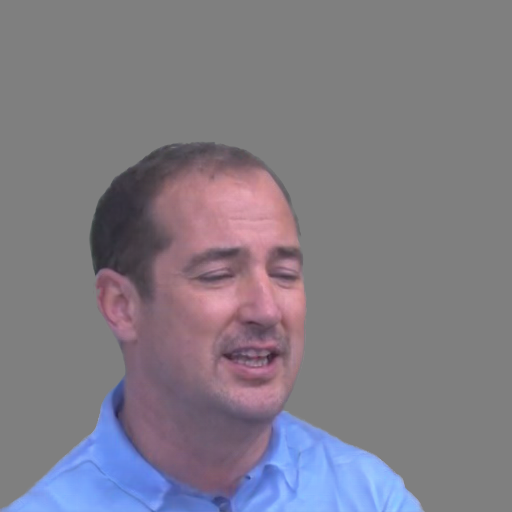}&
    \includegraphics[width=0.2\linewidth]{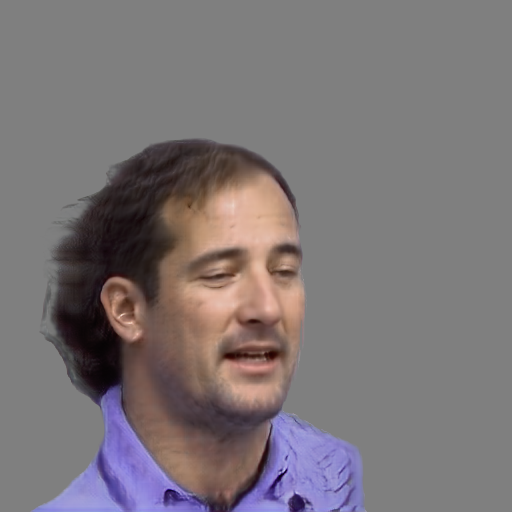}&
    \includegraphics[width=0.2\linewidth] {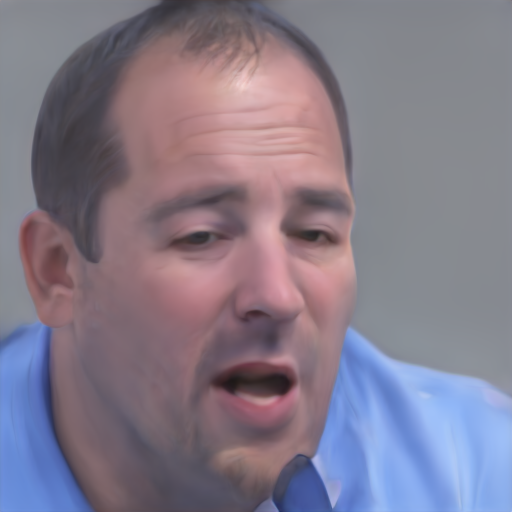}\\
    \includegraphics[width=0.2\linewidth]{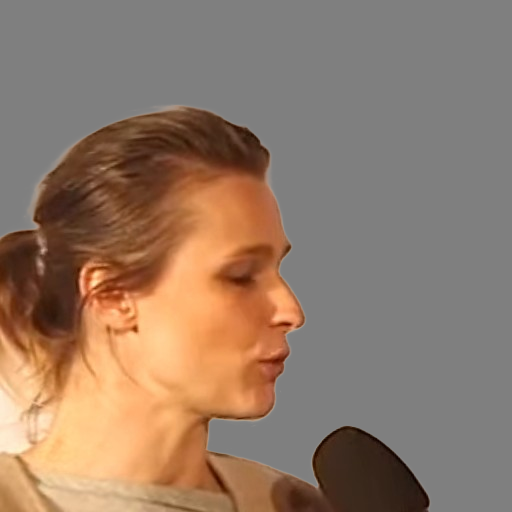}&
    \includegraphics[width=0.2\linewidth]{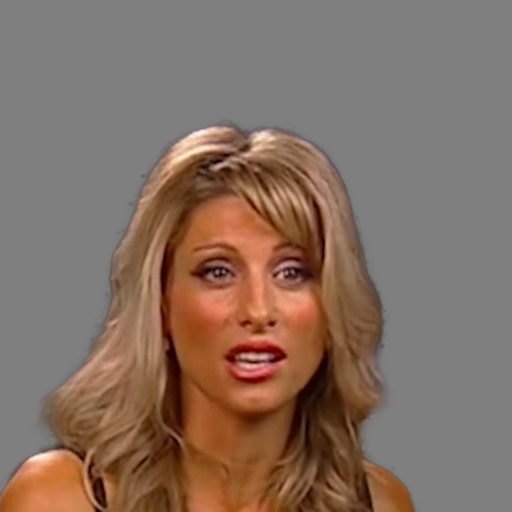}&
    \includegraphics[width=0.2\linewidth]{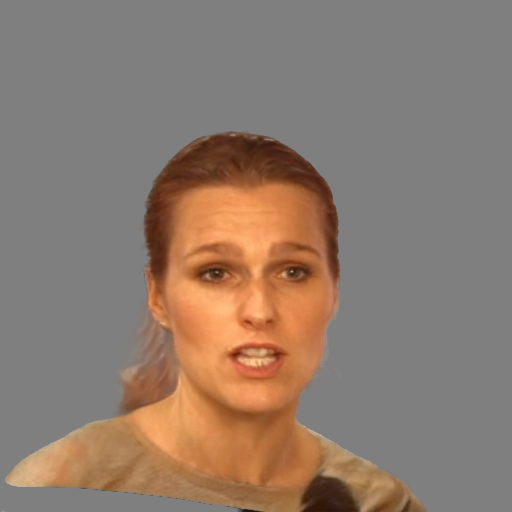}&
    \includegraphics[width=0.2\linewidth]{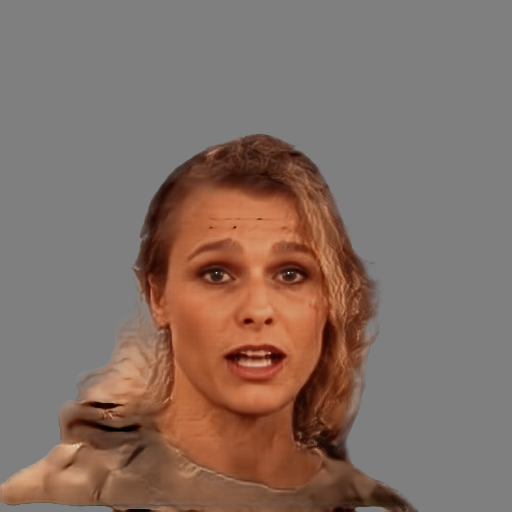}&
    \includegraphics[width=0.2\linewidth]{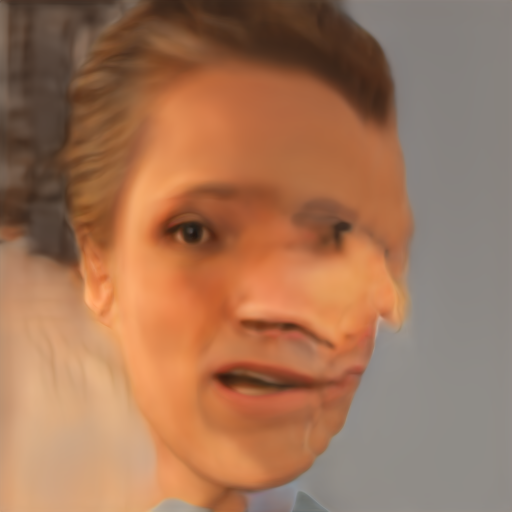}\\
    \includegraphics[width=0.2\linewidth]{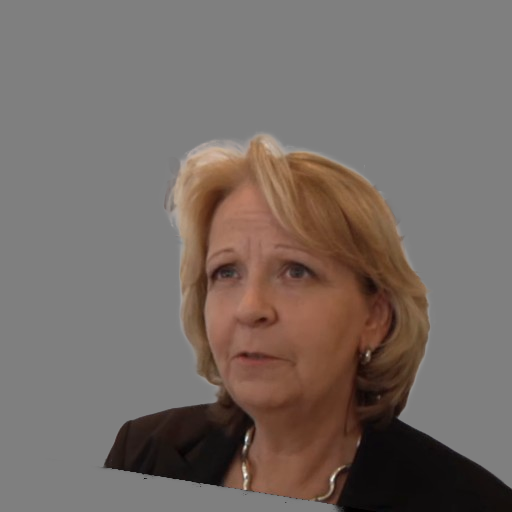}&
    \includegraphics[width=0.2\linewidth]{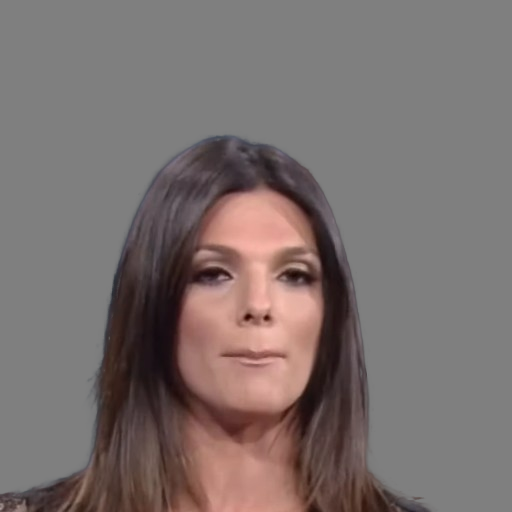}&
    \includegraphics[width=0.2\linewidth]{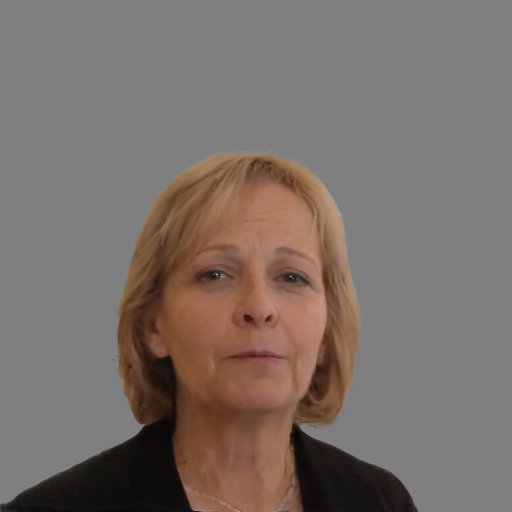}&
    \includegraphics[width=0.2\linewidth]{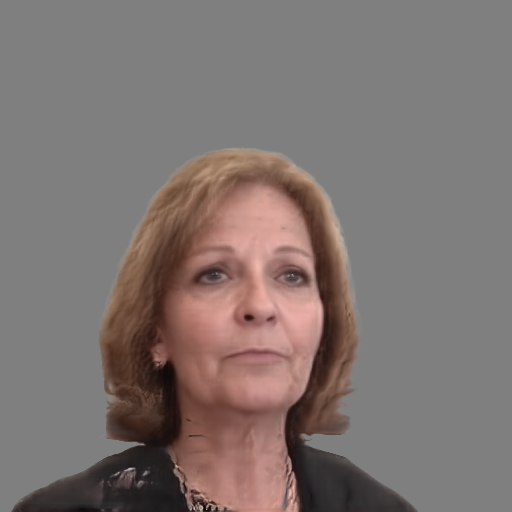}&
    \includegraphics[width=0.2\linewidth]{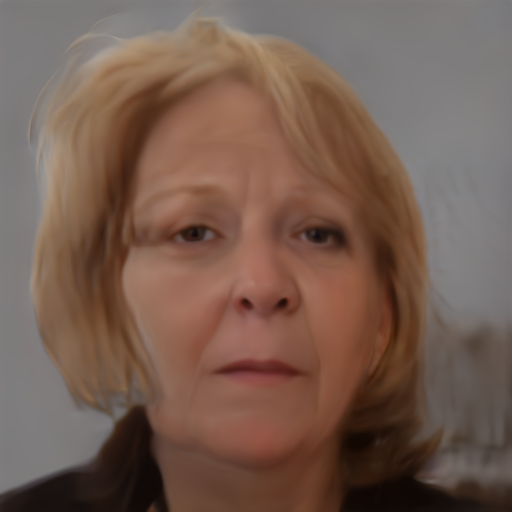}\\
    \includegraphics[width=0.2\linewidth]{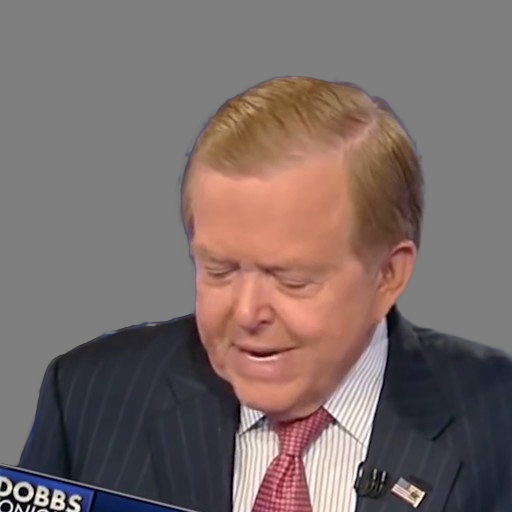}&
    \includegraphics[width=0.2\linewidth]{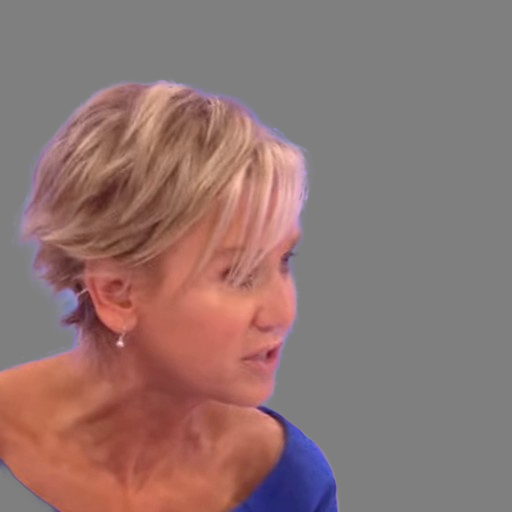}&
    \includegraphics[width=0.2\linewidth]{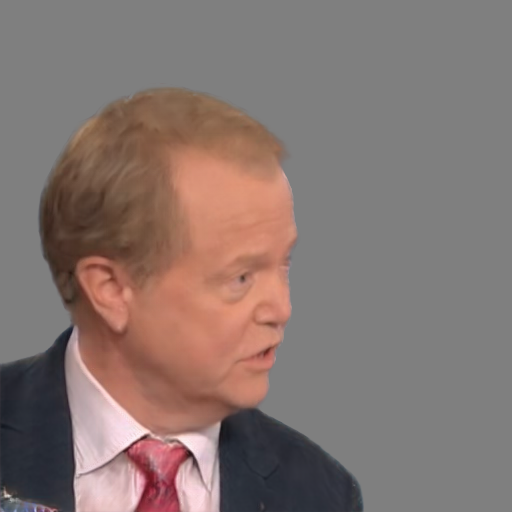}&
    \includegraphics[width=0.2\linewidth]{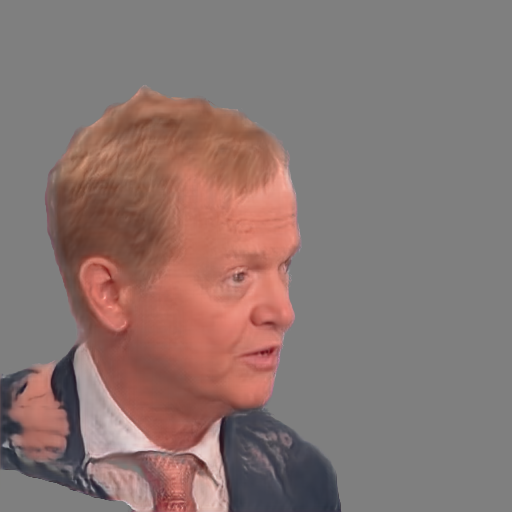}&
    \includegraphics[width=0.2\linewidth]{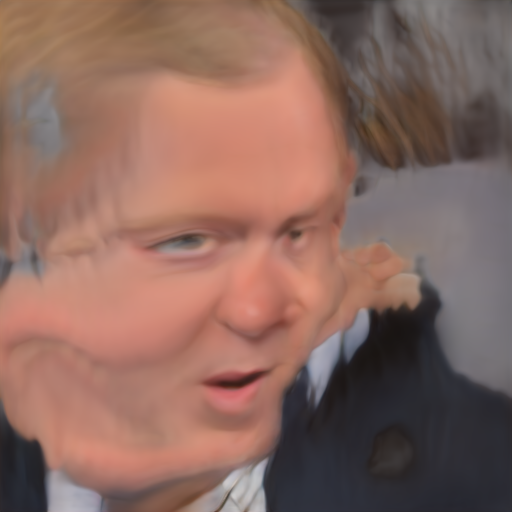}\\
  \end{tabular}
  \caption{Cross-reenactment results on $512 \times 512$} resolution
\end{figure*}

\begin{figure*}[t!]
  \centering
  \begin{tabular}{*{5}{@{\hspace{0pt}}c}}
  Source & Target & \small{GHOST 2.0} & HeSer \cite{shu2022few}& \tiny{StyleHEAT \cite{yin2022styleheat}}\\
  \includegraphics[width=0.2\linewidth]{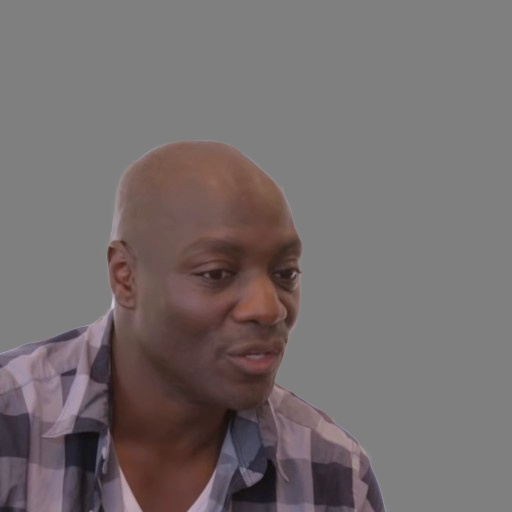}&
    \includegraphics[width=0.2\linewidth]{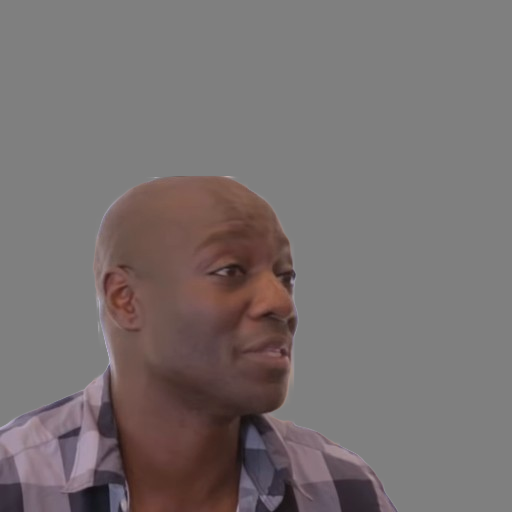}&
    \includegraphics[width=0.2\linewidth]{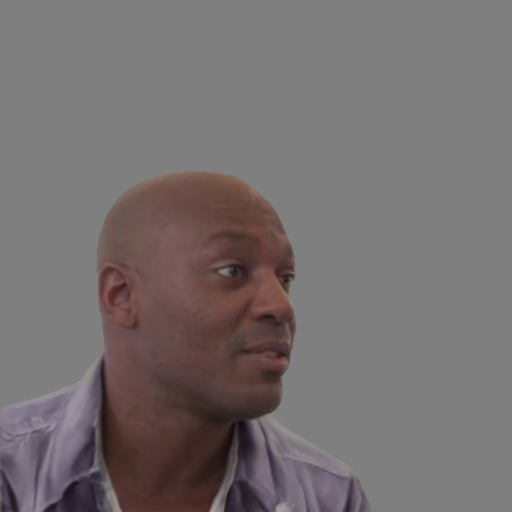}&
    \includegraphics[width=0.2\linewidth]{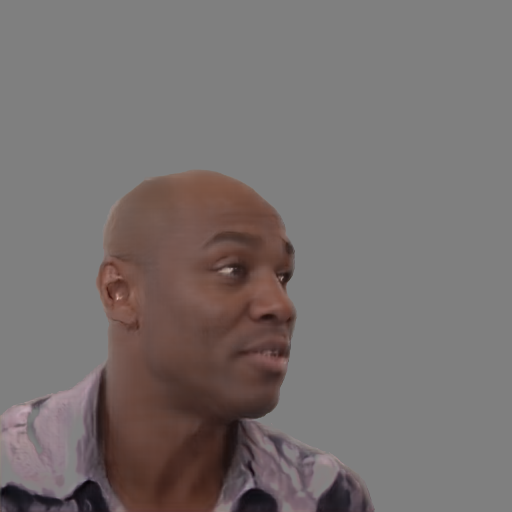}&
    \includegraphics[width=0.2\linewidth]{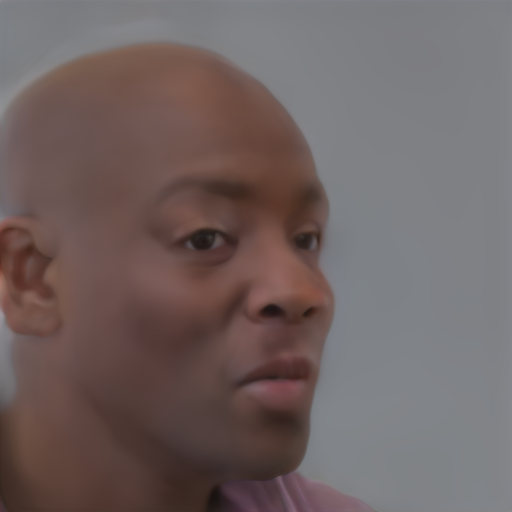}\\
    \includegraphics[width=0.2\linewidth]{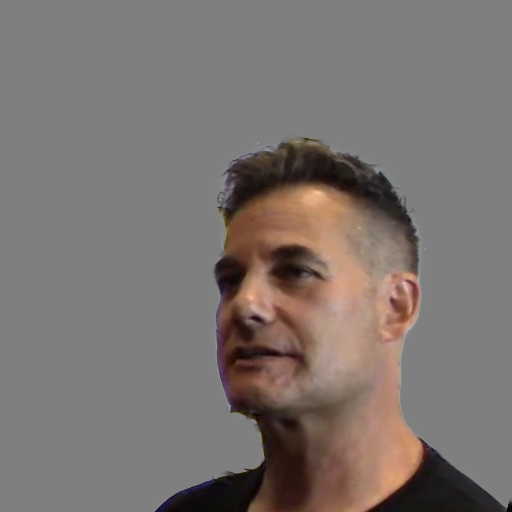}&
    \includegraphics[width=0.2\linewidth]{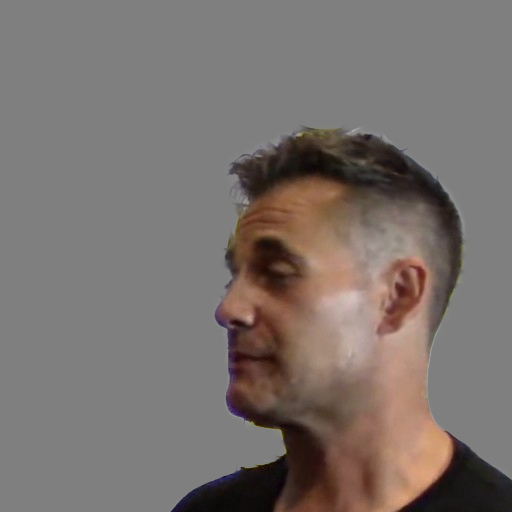}&
    \includegraphics[width=0.2\linewidth]{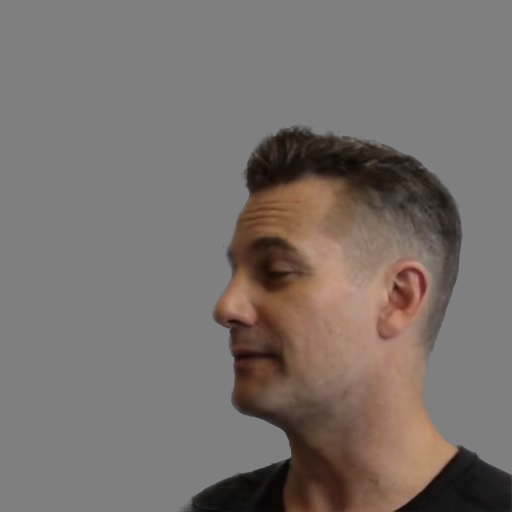}&
    \includegraphics[width=0.2\linewidth]{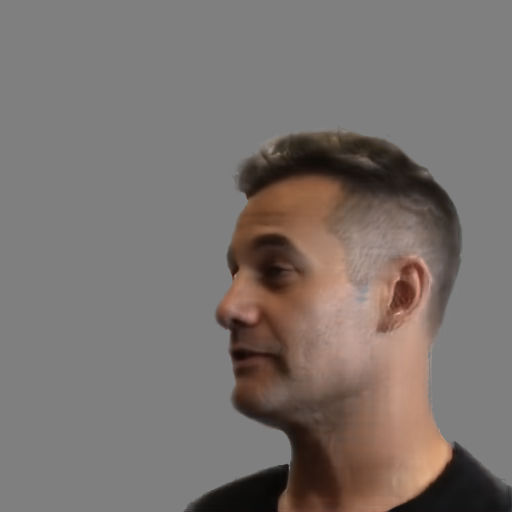}&
    \includegraphics[width=0.2\linewidth]{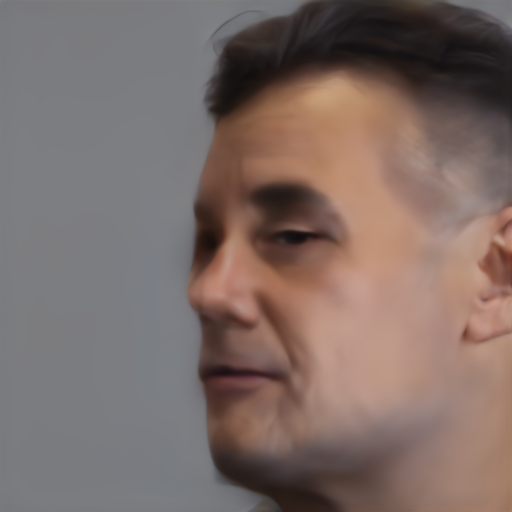}\\
    \includegraphics[width=0.2\linewidth]{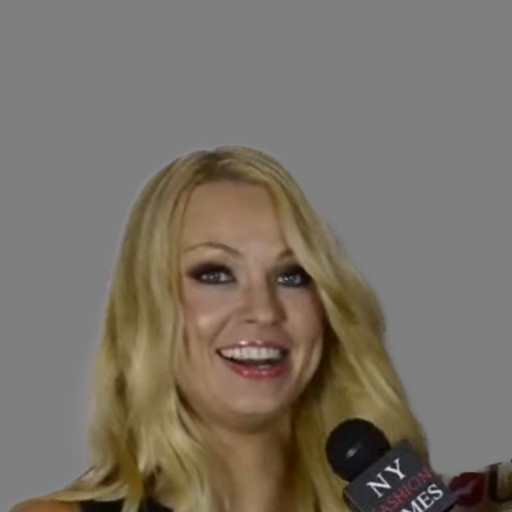}&
    \includegraphics[width=0.2\linewidth]{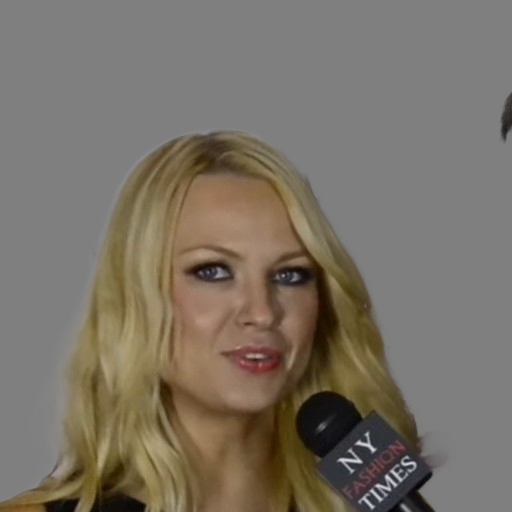}&
    \includegraphics[width=0.2\linewidth]{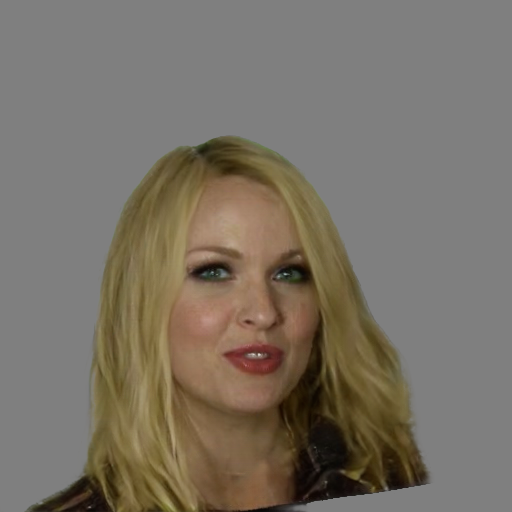}&
    \includegraphics[width=0.2\linewidth]{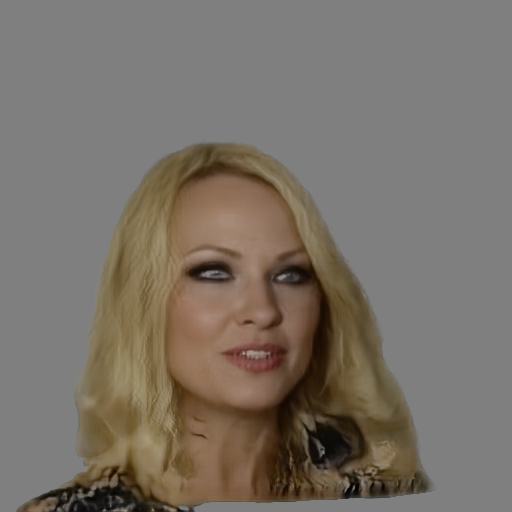}&
    \includegraphics[width=0.2\linewidth]{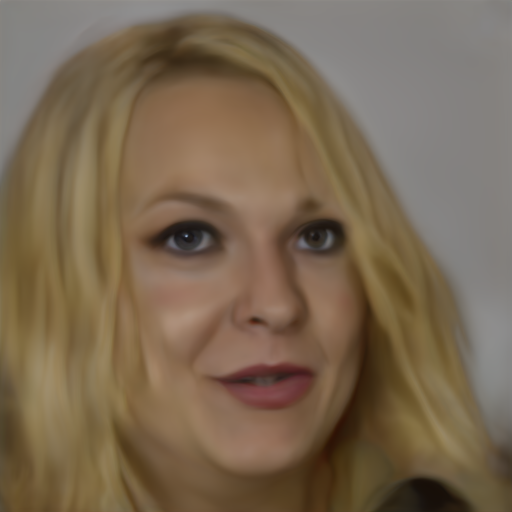}\\
    \includegraphics[width=0.2\linewidth]{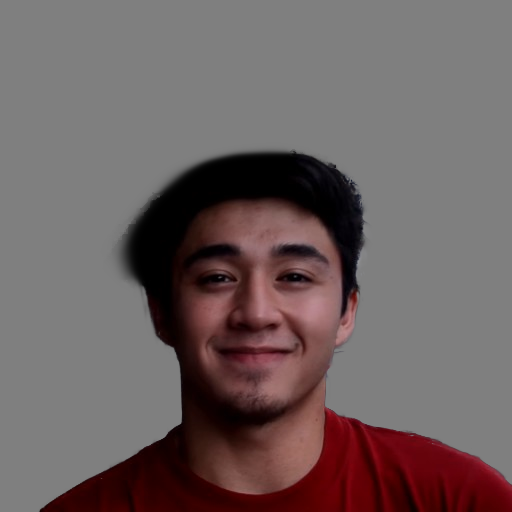}&
    \includegraphics[width=0.2\linewidth]{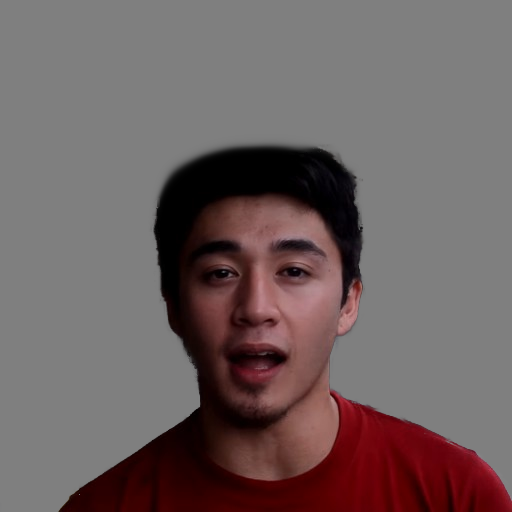}&
    \includegraphics[width=0.2\linewidth]{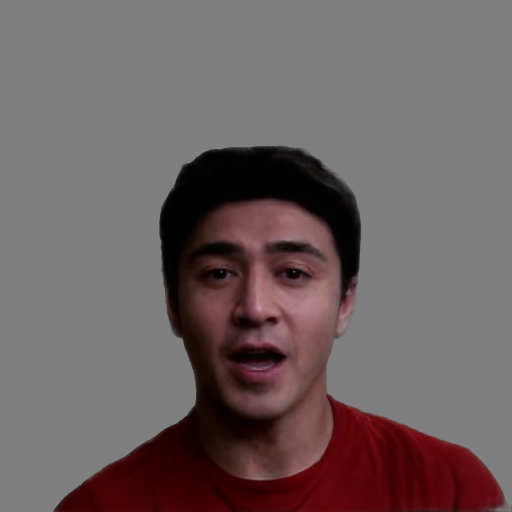}&
    \includegraphics[width=0.2\linewidth]{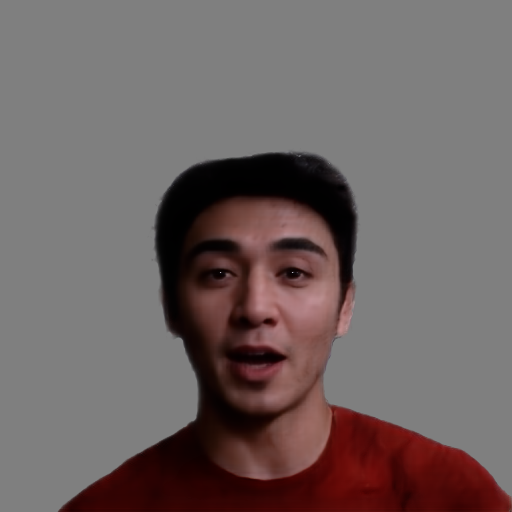}&
    \includegraphics[width=0.2\linewidth]{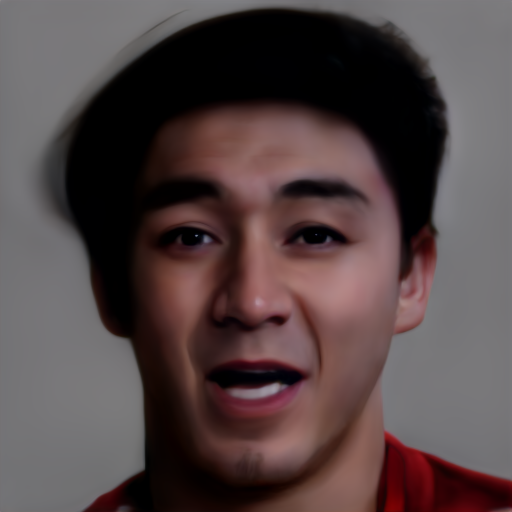}\\
    \includegraphics[width=0.2\linewidth]{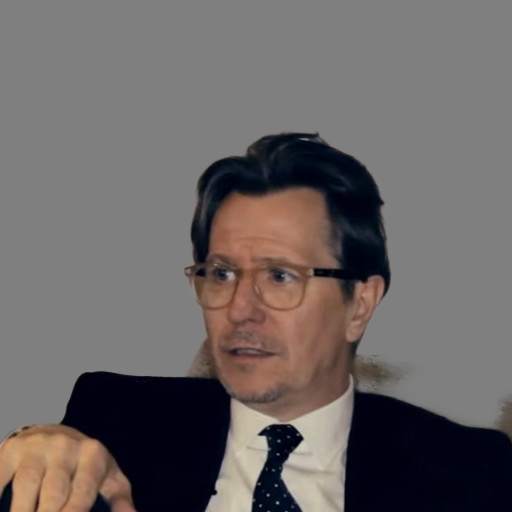}&
    \includegraphics[width=0.2\linewidth]{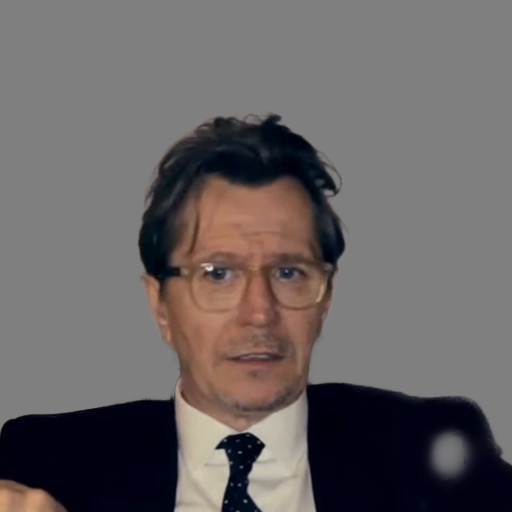}&
    \includegraphics[width=0.2\linewidth]{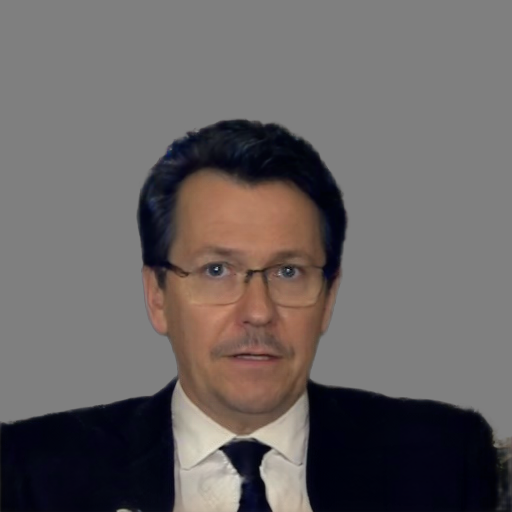}&
    \includegraphics[width=0.2\linewidth]{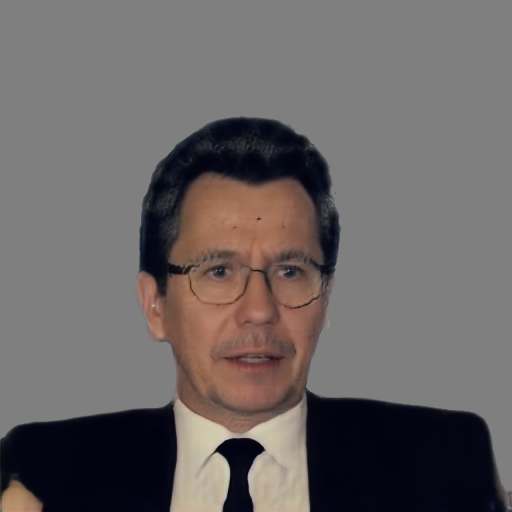}&
    \includegraphics[width=0.2\linewidth]{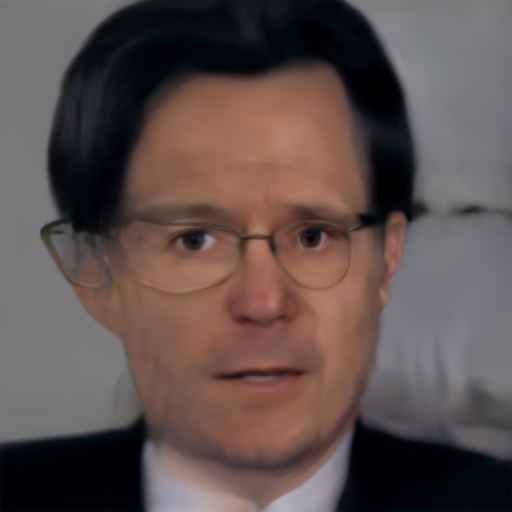}\\
    \includegraphics[width=0.2\linewidth]{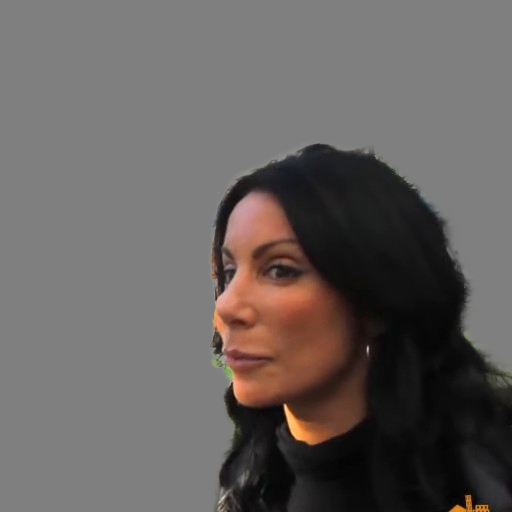}&
    \includegraphics[width=0.2\linewidth]{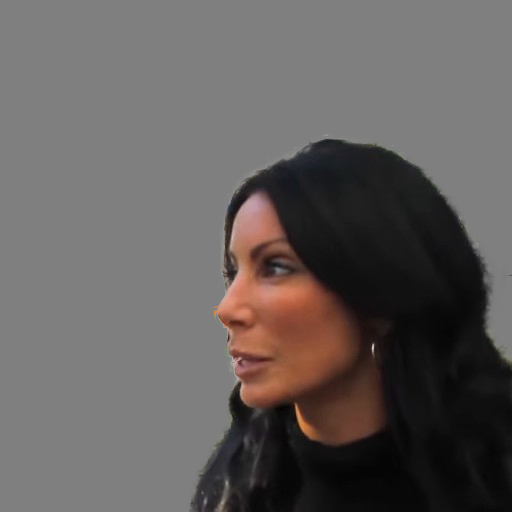}&
    \includegraphics[width=0.2\linewidth]{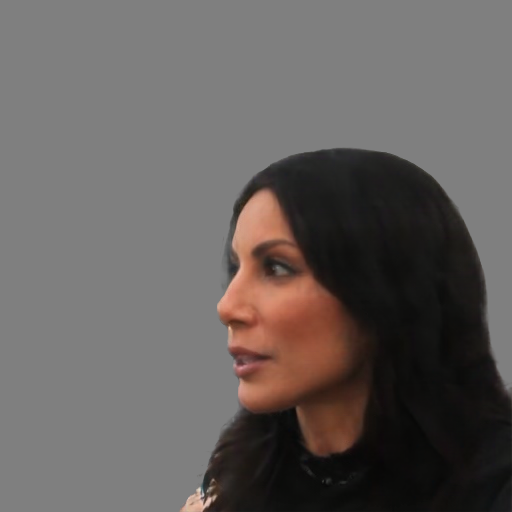}&
    \includegraphics[width=0.2\linewidth]{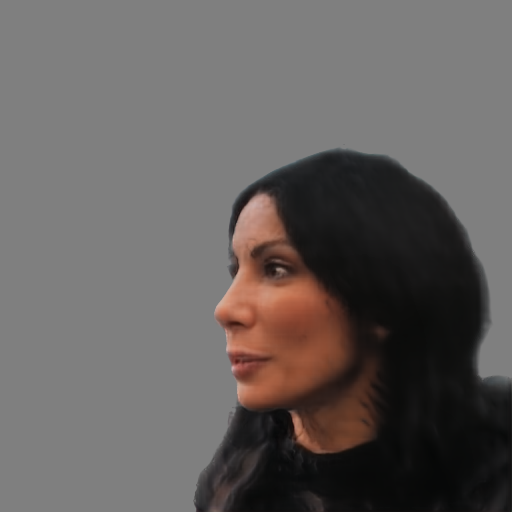}&
    \includegraphics[width=0.2\linewidth]{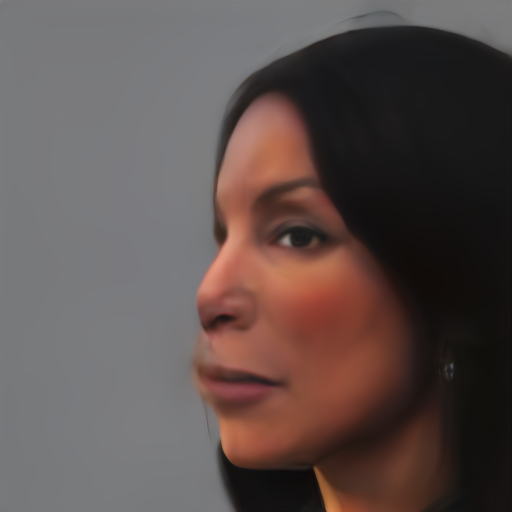}\\
  \end{tabular}
  \caption{Self-reenactment results on $512 \times 512$} resolution
\end{figure*}

\begin{figure*}[t!]
  \centering
  \begin{tabular}{*{7}{@{\hspace{0pt}}c}}
  Source & Target & GHOST 2.0 & DPE & DaGAN & LIA & TPSMM\\
  \includegraphics[width=0.14\linewidth]{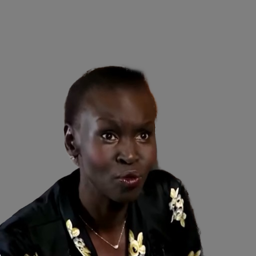}&
    \includegraphics[width=0.14\linewidth]{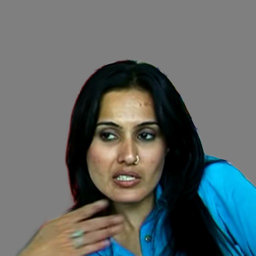}&
    \includegraphics[width=0.14\linewidth]{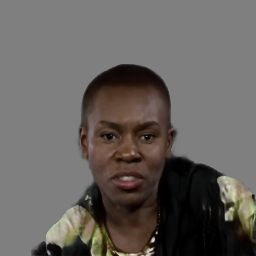}&
    \includegraphics[width=0.14\linewidth]{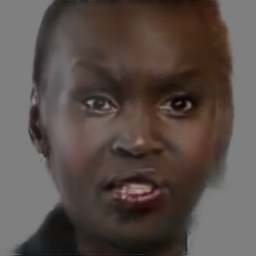}&
    \includegraphics[width=0.14\linewidth]{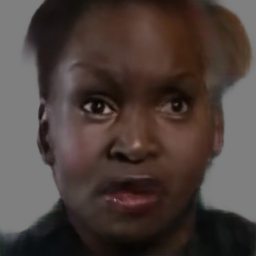}&
    \includegraphics[width=0.14\linewidth]{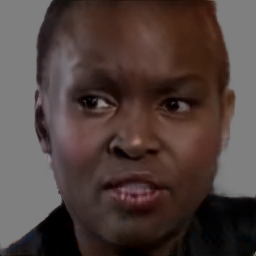}&
    \includegraphics[width=0.14\linewidth]{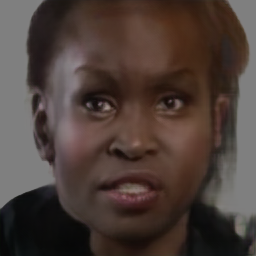}\\
    \includegraphics[width=0.14\linewidth]{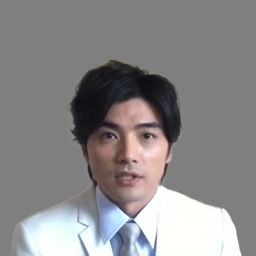}&
    \includegraphics[width=0.14\linewidth]{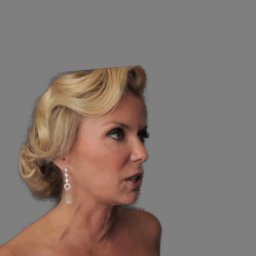}&
    \includegraphics[width=0.14\linewidth]{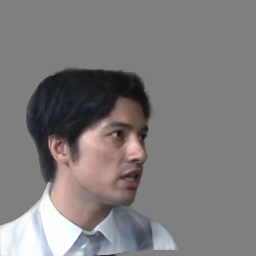}&
    \includegraphics[width=0.14\linewidth]{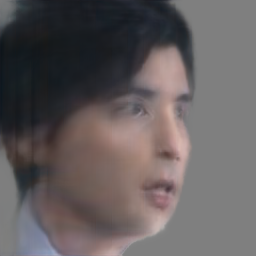}&
    \includegraphics[width=0.14\linewidth]{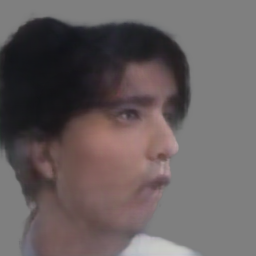}&
    \includegraphics[width=0.14\linewidth]{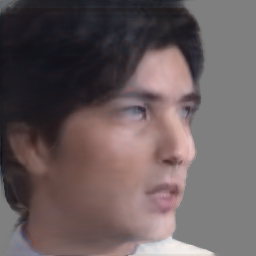}&
    \includegraphics[width=0.14\linewidth]{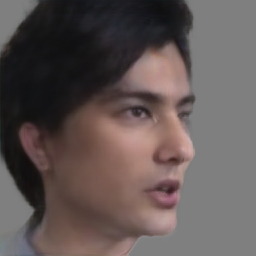}\\
    \includegraphics[width=0.14\linewidth]{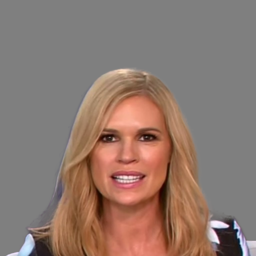}&
    \includegraphics[width=0.14\linewidth]{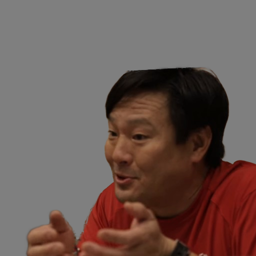}&
    \includegraphics[width=0.14\linewidth]{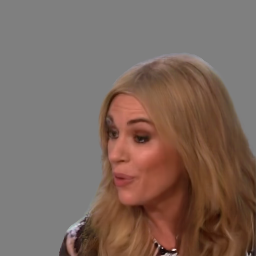}&
    \includegraphics[width=0.14\linewidth]{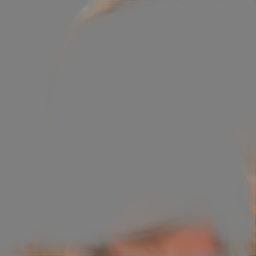}&
    \includegraphics[width=0.14\linewidth]{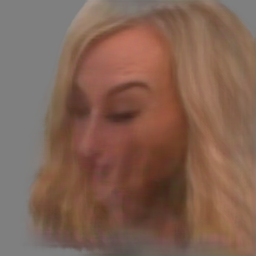}&
    \includegraphics[width=0.14\linewidth]{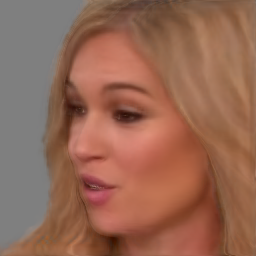}&
    \includegraphics[width=0.14\linewidth]{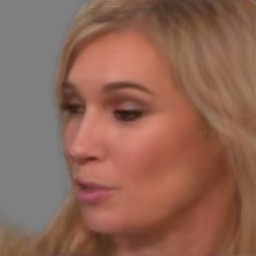}\\
    \includegraphics[width=0.14\linewidth]{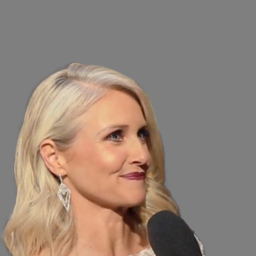}&
    \includegraphics[width=0.14\linewidth]{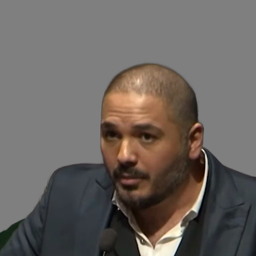}&
    \includegraphics[width=0.14\linewidth]{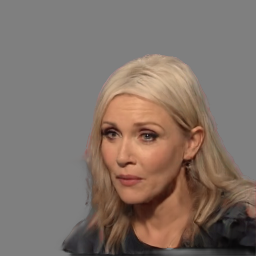}&
    \includegraphics[width=0.14\linewidth]{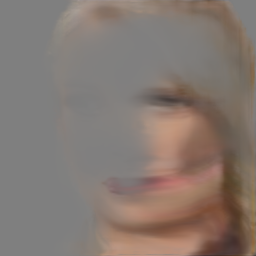}&
    \includegraphics[width=0.14\linewidth]{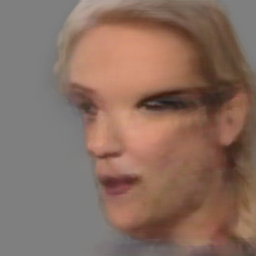}&
    \includegraphics[width=0.14\linewidth]{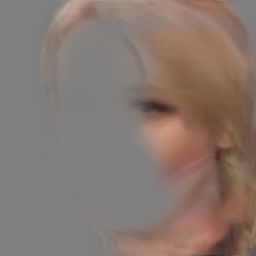}&
    \includegraphics[width=0.14\linewidth]{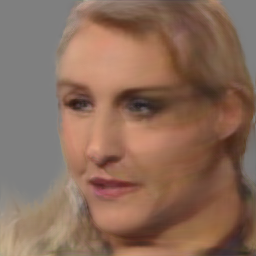}\\   
  \end{tabular}
\end{figure*}

\begin{figure*}[t!]
  \centering
  \begin{tabular}{*{7}{@{\hspace{0pt}}c}}
  Source & Target & GHOST 2.0 & X2face & FOMM & Bi-layer & PIRender\\
  \includegraphics[width=0.14\linewidth]{images/aligner_supp/cross_256_1/source_1.png}&
    \includegraphics[width=0.14\linewidth]{images/aligner_supp/cross_256_1/driver_1.png}&
    \includegraphics[width=0.14\linewidth]{images/aligner_supp/cross_256_1/our_1.png}&
    \includegraphics[width=0.14\linewidth]{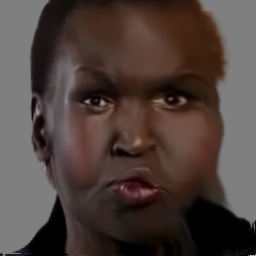}&
    \includegraphics[width=0.14\linewidth]{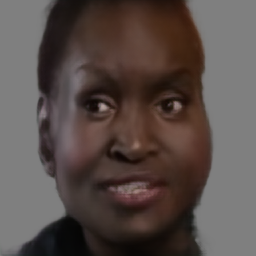}&
    \includegraphics[width=0.14\linewidth]{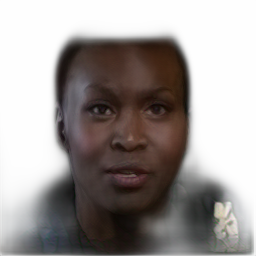}&
    \includegraphics[width=0.14\linewidth]{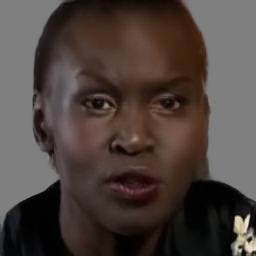}\\
     \includegraphics[width=0.14\linewidth]{images/aligner_supp/cross_256_1/source_3.png}&
    \includegraphics[width=0.14\linewidth]{images/aligner_supp/cross_256_1/driver_3.png}&
    \includegraphics[width=0.14\linewidth]{images/aligner_supp/cross_256_1/our_3.png}&
    \includegraphics[width=0.14\linewidth]{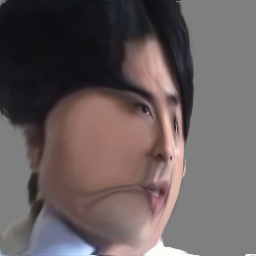}&
    \includegraphics[width=0.14\linewidth]{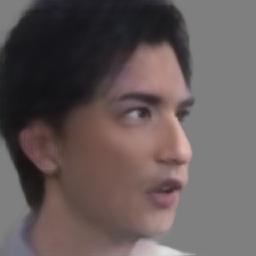}&
    \includegraphics[width=0.14\linewidth]{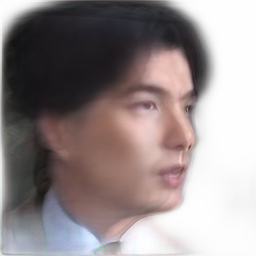}&
    \includegraphics[width=0.14\linewidth]{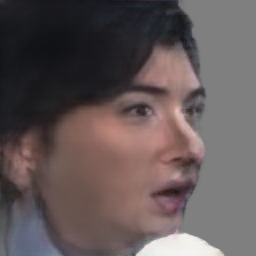}\\
     \includegraphics[width=0.14\linewidth]{images/aligner_supp/cross_256_1/source_4.png}&
    \includegraphics[width=0.14\linewidth]{images/aligner_supp/cross_256_1/driver_4.png}&
    \includegraphics[width=0.14\linewidth]{images/aligner_supp/cross_256_1/our_4.png}&
    \includegraphics[width=0.14\linewidth]{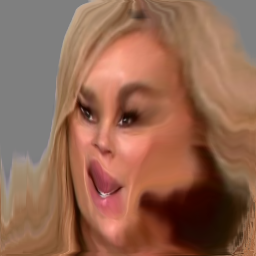}&
    \includegraphics[width=0.14\linewidth]{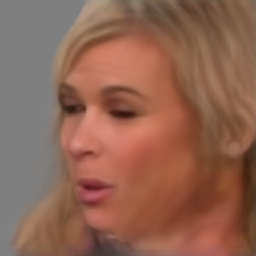}&
    \includegraphics[width=0.14\linewidth]{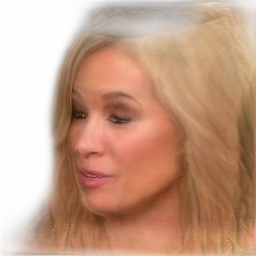}&
    \includegraphics[width=0.14\linewidth]{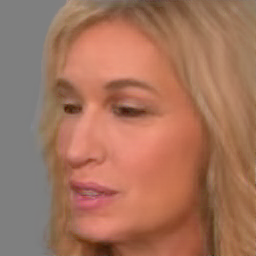}\\
     \includegraphics[width=0.14\linewidth]{images/aligner_supp/cross_256_1/source_2.png}&
    \includegraphics[width=0.14\linewidth]{images/aligner_supp/cross_256_1/driver_2.png}&
    \includegraphics[width=0.14\linewidth]{images/aligner_supp/cross_256_1/our_2.png}&
    \includegraphics[width=0.14\linewidth]{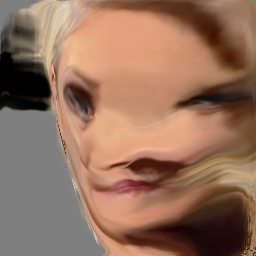}&
    \includegraphics[width=0.14\linewidth]{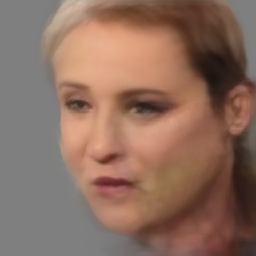}&
    \includegraphics[width=0.14\linewidth]{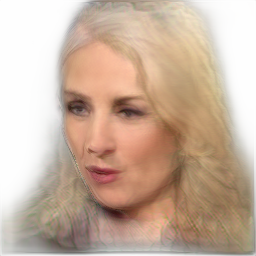}&
    \includegraphics[width=0.14\linewidth]{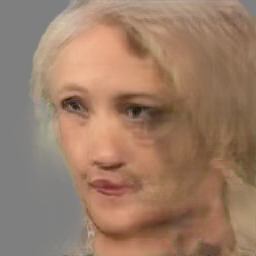}\\
      
  \end{tabular}
  \caption{Cross-reenactment results on $256 \times 256$} resolution
\end{figure*}

\begin{figure*}
\begin{center}
\centerline{\includegraphics[width=\textwidth]{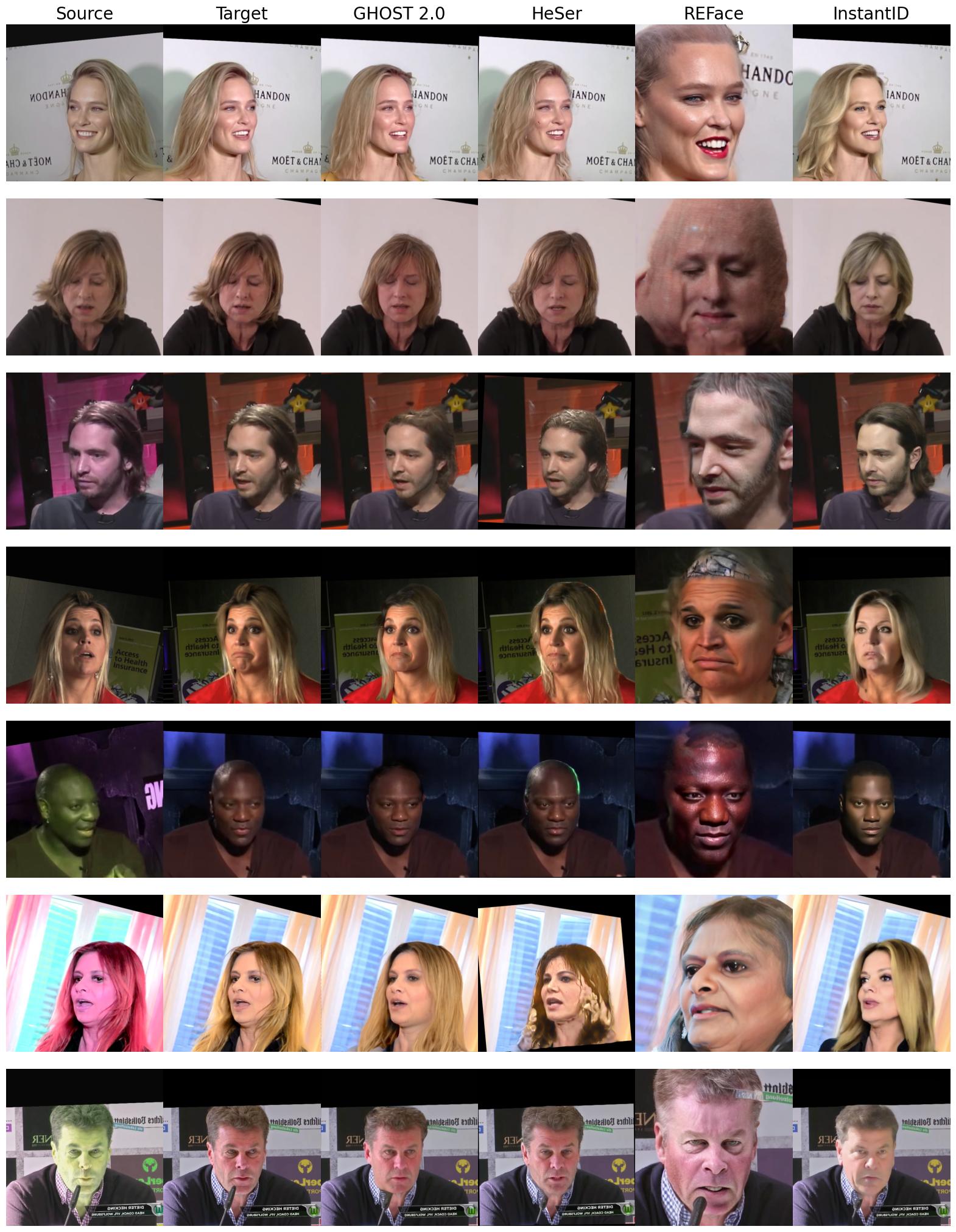}}
\caption{Head swap results with the same identity and color augmentation applied to the source}
\end{center}
\vskip -0.2in
\end{figure*}

\begin{figure*}
\begin{center}
\centerline{\includegraphics[width=\textwidth]{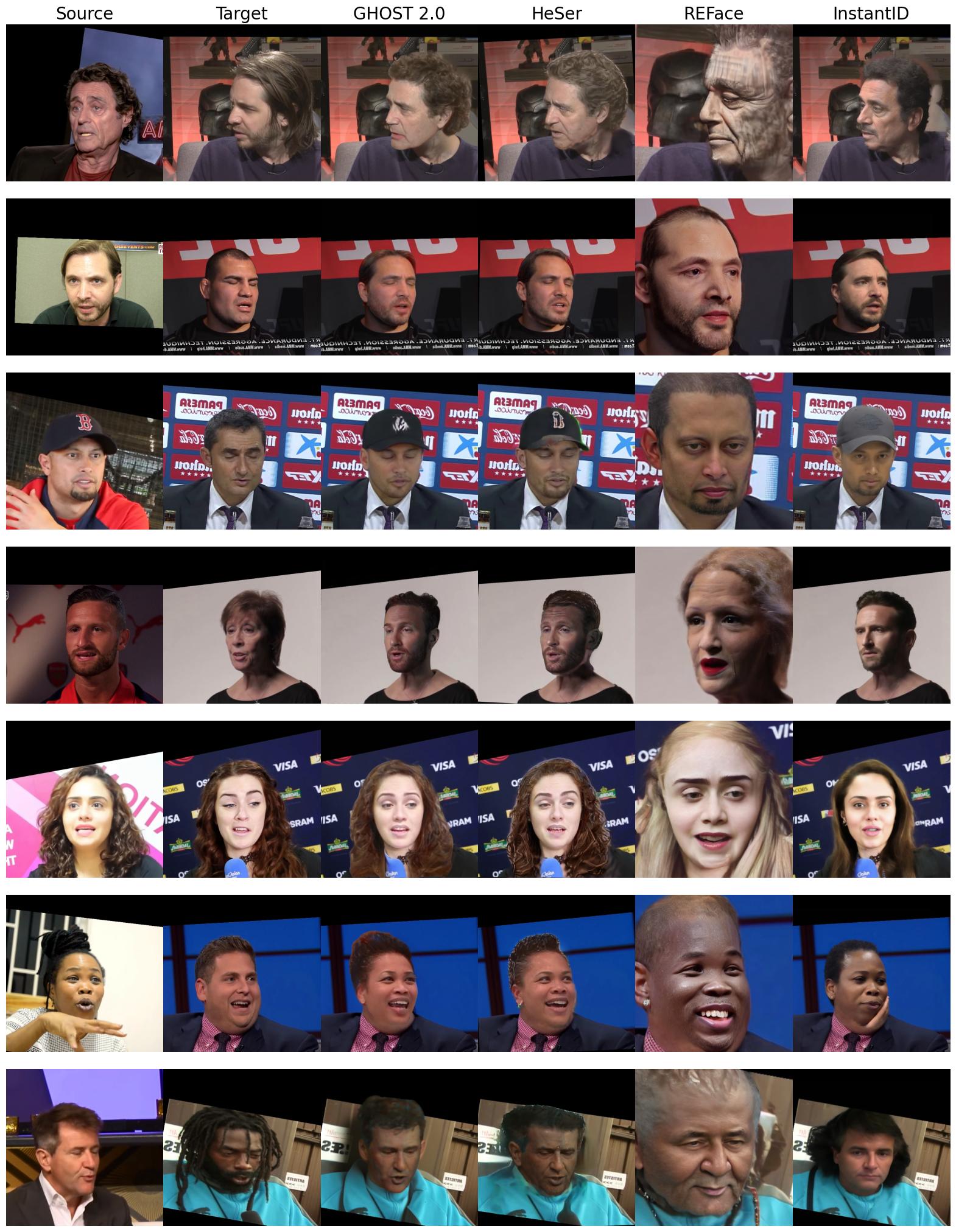}}
\caption{Head swap results with different identities}
\end{center}
\vskip -0.2in
\end{figure*}

\begin{figure*}
\begin{center}
\centerline{\includegraphics[width=\textwidth]{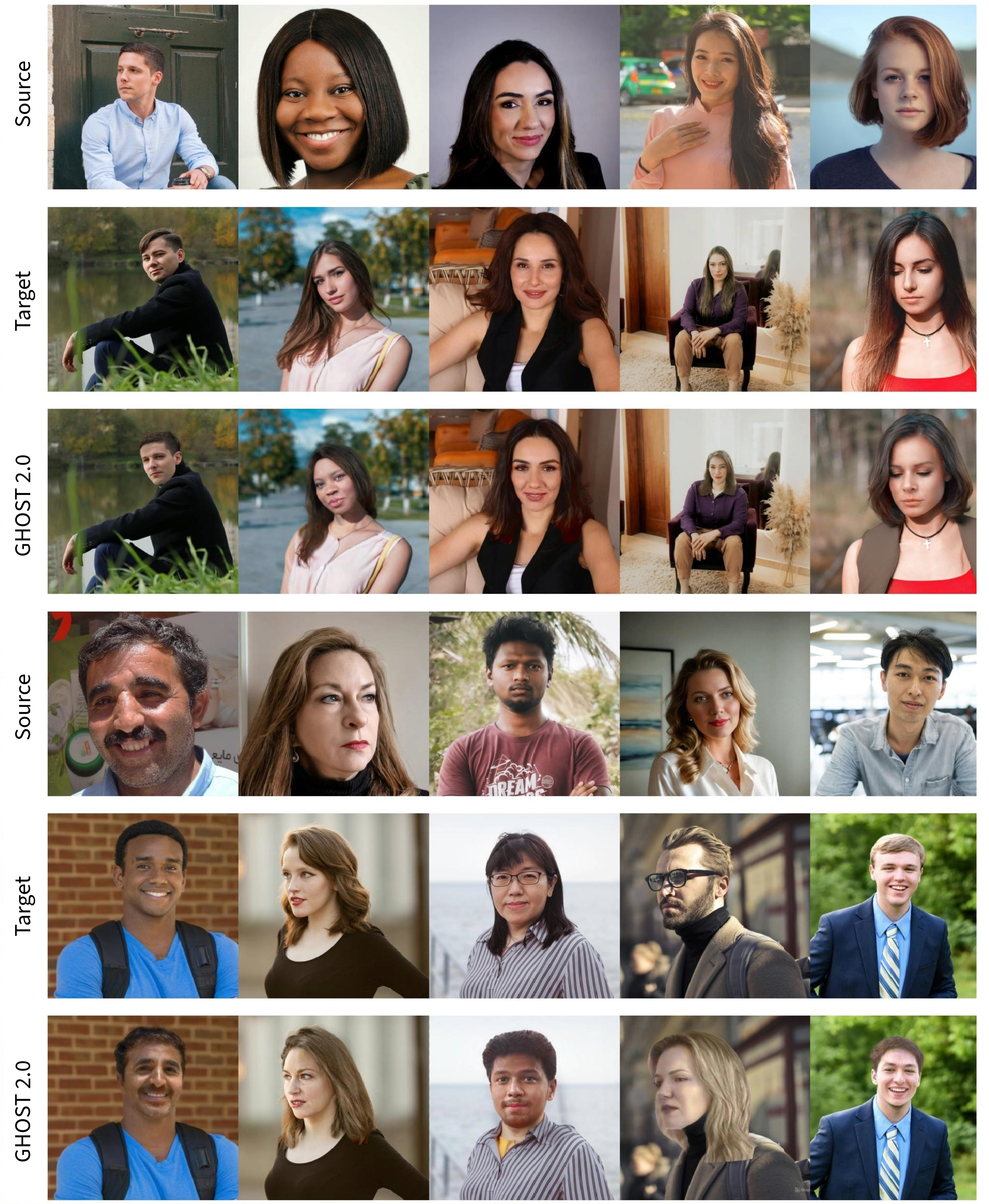}}
\caption{Head swap results on real-life and outdoor photos}
\label{real}
\end{center}
\vskip -0.2in
\end{figure*}

\end{document}